\documentclass[letterpaper, 10 pt, conference]{ieeeconf}  

\IEEEoverridecommandlockouts                              
\usepackage{graphics} 
\usepackage{epsfig} 
\usepackage{times} 

\usepackage{amsmath,amsthm,amssymb}
\usepackage{mathtools}
\usepackage{mathrsfs}
\usepackage{subfiles}
\usepackage{color}
\usepackage{enumerate}
\usepackage[caption=false]{subfig}

\newtheorem{theorem}{Theorem}

\newtheorem{proposition}[theorem]{Proposition}

\newtheorem{observation}[theorem]{Observation}
\newtheorem{example}[theorem]{Example}

\newcommand{\de}{\partial}
\newcommand{\R}{\mathbb{R}}

\newcommand{\tr}{^T}

\newcommand{\mc}{\mathcal}

\title{\LARGE \bf
An Optimal Task Allocation Strategy\\for Heterogeneous Multi-Robot Systems
}

\author{Gennaro Notomista$^{1}$, Siddharth Mayya$^{2}$, Seth Hutchinson$^{3}$, and Magnus Egerstedt$^{2}$
\thanks{This work was sponsored by ARL DCIST CRA W911NF-17-2-0181.}
\thanks{$^{1}$G. Notomista is with the School of Mechanical Engineering, Institute for Robotics and Intelligent Machines, Georgia Institute of Technology, Atlanta, GA, USA      {\tt\small g.notomista@gatech.edu}}%
\thanks{$^{2}$S. Mayya and M. Egerstedt are with the School of Electrical and Computer Engineering, Georgia Institute of Technology, Atlanta, GA, USA {\tt\small \{siddharth.mayya,magnus\}@gatech.edu}}%
\thanks{$^{3}$S. Hutchinson is with the School of Interactive Computing, Georgia Institute of Technology, Atlanta, GA, USA  {\tt\small seth@gatech.edu}}%
}
\begin{document}

\maketitle
\thispagestyle{empty}
\pagestyle{empty}

\begin{abstract}

For a team of heterogeneous robots executing multiple tasks, we propose a novel algorithm to optimally allocate tasks to robots while accounting for their different capabilities. Motivated by the need that robot teams have in many real-world applications of remaining operational for long periods of time, we allow each robot to choose tasks taking into account the energy consumed by executing them, besides the global specifications on the task allocation. The tasks are encoded as constraints in an energy minimization problem solved at each point in time by each robot. The \textit{prioritization} of a task over others -- effectively signifying the allocation of the task to that particular robot -- occurs via the introduction of slack variables in the task constraints. Moreover, the suitabilities of certain robots towards certain tasks are also taken into account to generate a task allocation algorithm for a team of robots with heterogeneous capabilities. The efficacy of the developed approach is demonstrated both in simulation and on a team of real robots.

\end{abstract}

\section{INTRODUCTION}
\label{sec:introduction}
Multi-robot systems exhibit desirable reconfigurability and robustness properties, which make them suitable for executing a wide range of tasks \cite{yang2018grand}. Indeed, the overall goal of their deployment often consists of executing more than one task \cite{khamis2015multi}. For example, the robots deployed in a disaster scenario might need to perform environment exploration, source-seeking, as well as object manipulation (see e.g. \cite{scerri2005allocating}). \par 

A natural question arising in this context is: which task should be assigned to which robot? In fact, \textit{task allocation} is a widely studied topic in multi-robot systems, see e.g. the surveys \cite{gerkey2004formal,dias2006market}. Different algorithms have been proposed, which account for factors such as resource or time constraints \cite{luo2013distributed}, limited energy availability for the robots \cite{wu2018gini}, as well as communication topologies among the robots \cite{choi2009consensus}. Differently from typical task allocation algorithms, which result in each robot executing either one or a subset of the tasks, in this paper, we allow robots to execute all the tasks at the same time with different \textit{priorities}. The task allocation is then effectively realized through a different task priority assignment among the robots. \par 

This paper focuses on a specific class of multi-robot tasks which can be encoded via a cost, that is function of the state of the system. The execution of the task is identified with the minimization of the cost, whose value is inversely proportional to the extent to which the task has been accomplished. Such a description of multi-robot tasks has been used to generate a wide variety of behaviors, such as environment surveillance and exploration, formation-constrained control and path following \cite{cortes2017coordinated}. \par 

Typically, the execution of multi-robot tasks such as the ones mentioned above might require the robots to operate in real world environments for extended periods of time. When designing a task allocation strategy, it is therefore desirable to impose \emph{survivability} constraints \cite{egerstedt2018robot}, which would allow robots to operate in uncertain and changing environmental conditions for long periods of time under limited energy resources. Motivated by this idea, we develop an optimization-based task allocation strategy which is capable of explicitly taking into account the energy that the robots would spend to execute the assigned tasks. \par

Moreover, in many applications, robots in a multi-robot team are seldom identical \cite{balch2002robot}: they might be equipped with different sensory and actuation suites or differ from each other in the available energy (due to varying battery levels) and the extent of wear and tear in the hardware \cite{parker2000lifelong}. Such \emph{heterogeneity} among the robots affects their ability to perform different tasks, and an effective algorithm to allocate tasks among robots should take into account their suitability for each given task. We thus develop a task allocation algorithm which explicitly accounts for the heterogeneity in the suitability of robots for different tasks, as well as the survivability constraint mentioned above.

In the context of long-term autonomy, offline optimal task allocation routines, although computationally not intensive for the robots, suffer from the fragility typical of optimal control strategies \cite{egerstedt2018robot}. Consequently, this paper presents a dynamic task allocation algorithm, formulated as an optimization problem which is efficient enough to be solved by the robots at each point in time. For a given robot, a particular way to ensure the execution of $M$ tasks while taking into account survivability considerations, is to solve the following optimization problem at each point in time:
\begin{align*}
\min_{u} & ~~ \|u\|^2 \\
\text{s.t.} & ~~c_{task_i}(x,u) \geq 0, \forall i \in \{1,\ldots,M\},
\end{align*}
where $u$ is the control effort expended by the robot, $x$ is its state, and $c_{task_i}$ denotes a constraint function which ensures the execution of task $i$. Such a \emph{constraint-based} formulation allows for higher flexibility and robustness when compared to purely cost-based optimization problems, especially in the context of long-term autonomy applications \cite{egerstedt2018robot}. \par 

In order to allow such an optimization problem to remain feasible during the execution of multiple tasks by the robots, each task constraint is augmented with a slack variable corresponding to the effectiveness of performing that task:
\begin{align*}
\min_{u,\delta} & ~~ \|u\|^2 + \|\delta\|^2 \\
\text{s.t.} & ~~c_{task_i}(x,u) \geq - \delta_i, \forall i \in \{1,\ldots,M\}.
\end{align*}
We illustrate that not only do the slack variables $\delta = [\delta_1,\ldots,\delta_M]^T$ enable the feasibility of such an optimization program, they allow individual robots to prioritize tasks, i.\,e. perform some tasks more effectively than others. Such a task prioritization can be embedded by adding additional constraints on the slack variables pertaining to each task, written as $K \delta \geq 0$. 
Here, $K$ is a matrix which can encode pairwise inequality constraints between the elements of the vector $\delta$. We demonstrate that such a formulation allows individual robots to perform the tasks with varying levels of priority, while taking into account long-term survivability, heterogeneity in their capabilities, and global requirements on the desired task allocation.

The outline of the paper is as follows. Section \ref{sec:lit} discusses relevant results from the multi-robot task allocation literature and compares them with the approach we propose in this paper. Then, results from non-linear control theory and our previous work \cite{arxiv:extendedacc} are presented, which will be used throughout the paper. Section \ref{sec:task_comp} formulates an optimization problem to compose multiple tasks as constraints within a single optimization problem. In Section \ref{sec:task_all}, we allow the robots to prioritize certain tasks by imposing constraints on the effectiveness with which different tasks must be performed, leading to the development of the task allocation framework. In Section \ref{sec:exp}, we show the results of the deployment of the task allocation algorithm on a team of robots. Section \ref{sec:conc} concludes the paper.

\section{BACKGROUND AND RELATED WORK} \label{sec:lit}
\subsection{Literature Review}
Task allocation is an extensively studied topic in the multi-robot systems literature (see, for instance, the survey and taxonomy papers \cite{khamis2015multi,gerkey2004formal,korsah2013comprehensive} and the references within). Many different algorithms have been developed to allocate tasks to robots in a team, e.\,g. auction-based approaches \cite{bertsekas1990auction,dias2004traderbots}, distributed assignment algorithms \cite{giordani2010distributed} and stochastic methods \cite{berman2009optimized}. \par 

In addition to the development of efficient algorithms for task allocation, different strategies have been tailored for a number of application scenarios. For example, in \cite{tolmidis2013multi} and \cite{lemaire2004distributed}, the dynamic and distributed aspects of the task allocation problem are considered, respectively. In \cite{tang2007complete}, the authors consider a layered mechanism to perform task allocation in heterogeneous teams of robots. In \cite{luo2013distributed}, additional deadline constraints on the tasks that need to be performed are taken into account. \par  

In contrast to many task allocation approaches which assign one task at a time per robot, this paper studies a scenario where individual robots perform multiple tasks simultaneously with different priorities. We formulate this problem as a quadratic program (QP) which is solved at each point in time. Such a formulation has a lower computational complexity \cite{boyd2004convex} when compared to other algorithms which assign multiple tasks to robots \cite{gerkey2004formal}. We demonstrate that, solving the optimization problem at each instant in time leads to a dynamic task allocation algorithm which can adapt to changing allocation requirements. Furthermore, such an optimization framework allows us to encode the heterogeneity of robots and to develop a task allocation mechanism which minimizes the control effort expended by the robots. \par 

The next section introduces some concepts from non-linear control theory and from our previous work \cite{arxiv:extendedacc} which will be used throughout the paper.

\subsection{Constraint-Based Task Execution}
\label{subsec:cbof}
Given a continuously differentiable function $h:\R^n\to\R$, define the \textit{safe set} $S$ as its zero-superlevel set:
\begin{equation}\label{eq:safeset}
	S = \{ x\in \R^n~|~h(x)\ge0 \}.
\end{equation}
Let $\partial S = \{ x\in \R^n~|~h(x)=0 \}$ and $S^\circ = \{ x\in \R^n~|~h(x)>0 \}$ denote the boundary and the interior of $S$, respectively. The function $h$ is called a \textit{(zeroing) control barrier function (ZCBF)} if the following condition is satisfied:
\begin{equation}\label{eq:uzcbf}
	\sup_{u\in U} \left\{L_f h(x) + L_g h(x) u + \gamma(h(x))\right\} \ge 0\quad\forall x\in \R^n,
\end{equation}
where $\gamma$ is an extended class $\mc K$ function \cite{xu2015robustness}, and $L_f h(x)$ and $L_g h(x)$ denote the Lie derivatives of $h$ in the directions $f$ and $g$, respectively. The following theorem summarizes two important properties of ZCBFs.

\begin{theorem}\label{thm:zcbfproperties}
	Given a dynamical system in control affine form $\dot x = f(x) + g(x) u$, where $x\in\R^n$ and $u\in\R^m$ denote the state and the input, respectively, $f$ and $g$ are locally Lipschitz, and a set $S\subset\R^n$ defined by a continuously differentiable function $h$ as in \eqref{eq:safeset}, any Lipschitz continuous controller $u$ such that \eqref{eq:uzcbf} holds renders the set $S$ forward invariant and asymptotically stable, i.\,e.,:
	\begin{align*}
	&x(0)\in S \Rightarrow x(t)\in S~\forall t\ge0\\
	&x(0)\notin S \Rightarrow x(t)\rightarrow\in S~\text{as}~t\to\infty,
	\end{align*}
	where $x(0)$ denotes the state $x$ at time $t=0$.
\end{theorem}
\begin{proof}
See \cite{xu2015robustness} and \cite{arxiv:extendedacc}.
\end{proof}

As our primary objective is the allocation of tasks among different robots, we abstract the motion of the robots using single integrator dynamics, assuming that we can manipulate their velocities directly. Moreover, as discussed in Section \ref{sec:introduction}, we consider tasks that can be encoded by means of a positive and continuously differentiable cost function $J:\R^n\to\R$.

We are interested in synthesizing a control signal $u(t)$ that allows the minimization of the cost $J(x(t))$. This can be achieved by solving, at each point in time, the minimization problem
\begin{equation}\label{eq:minuJ}
\min_{u} J(x),
\end{equation}
where $x$ and $u$ are coupled through the single integrator dynamics $\dot x = u$.

As explained in \cite{egerstedt2018robot} and discussed in Section~\ref{sec:introduction}, the constraint-driven control strategy has advantages in terms of robustness against unpredictable and changing environmental conditions -- properties which are useful when considering long-duration autonomy. In \cite{arxiv:extendedacc}, we show that solving \eqref{eq:minuJ} in order to synthesize $u(t)$ is equivalent to solving the following constraint-based optimization problem, in the sense that they both achieve the goal of minimizing the cost $J$:
\begin{equation} 
\begin{aligned} \label{eqn_const_opt} 
\min_{u,\delta} & ~~\|u\|^2 + |\delta|^2 \\
\text{s.t.} & ~~ \frac{\partial h}{\partial x}u \geq -\gamma(h(x)) - \delta
\end{aligned}
\end{equation}
where $\delta \in \mathbb{R}$ is the slack variable signifying the extent to which the task constraint can be violated, $\gamma$ is an extended class $\mc K$ function, and $h(x) = -J(x)$ is a \emph{(zeroing) control barrier function}. The zero-superlevel set of $h$ is $S = \left\{x~|~h(x)\ge0\right\}= \left\{x~|~J(x)\le0\right\}= \left\{x~|~J(x)=0\right\}$, where the last equality holds because the cost $J(x)$ is a non-negative function. In the particular case in which $J$ is strictly convex and $J(0)=0$, we have that $\frac{\de J}{\de x}(x)\neq0~\forall x\neq0$. Then, Theorem~\ref{thm:zcbfproperties} directly implies that $x\to\in S$, i.\,e. $J(x(t))\to0$, as $t\to\infty$. For a proof of the general case, we refer to \cite{arxiv:extendedacc}.

In the next section, we introduce the idea of simultaneous execution of multiple tasks by each robot in a multi-robot team. This concept will be then used in Section~\ref{sec:task_all} in order to formulate an optimization problem which can be solved efficiently and through which the robots can \textit{automatically} prioritize tasks according to their heterogeneous capabilities and global specifications on the task allocation.

\section{Constraint-Based Multi-Task Execution} \label{sec:task_comp}
Consider a team of $N$ mobile robots operating in a compact domain $\mathcal{D}$. The robots need to execute $M$ different tasks denoted as $T_1,\ldots,T_M$. Let the index set of the tasks be denoted as $\mc M = \{1,\ldots,M\}$. As before, we assume that each task $T_m$ can be encoded as the minimization of a cost function $J_m$, $m \in \mc M$, and each robot is modeled as a single integrator, $\dot x_i=u_i,~i\in\mathcal{N}=\{1,\ldots,N\}$, $x_i\in\mathbb R^d$.

\begin{observation}
Note that, as the tasks are identified with a cost function $J$ introduced in Section~\ref{subsec:cbof}, they do not explicitly depend on time, i.\,e. $\frac{\de J_m}{\de t} = 0~,\forall m\in \mc M$.
\end{observation}

As defined in \eqref{eqn_const_opt}, the constraint-based optimization problem for a given robot $i$ performing $M$ tasks is given as,
\begin{equation}\label{eq:udeltaqp}
\begin{aligned}
\min_{u_i,\delta_i} &~~\|u_i\|^2 + l\|\delta_i\|^2 \\ 
 \text{s.t.} ~~ &\frac{\partial h_{m}}{\partial x_i} u_i \geq -\gamma(h_{m}(x)) - \delta_{i,m}, ~ \forall m \in \mc M\\
& \|\delta_{i}\|_\infty \leq \delta_{max}
\end{aligned}
\end{equation}
where $l\ge0$ is a scaling constant, $\delta_i = [\delta_{i,1},\ldots,\delta_{i,M}]\tr$, $h_{m}(x) = -J_{m}(x)$, and $\delta_{max}$ signifies the maximum allowable value that each task constraint can be relaxed\footnote{Note that, since the robots move in a compact domain and $J_m\in C^1$, $J_m$ is bounded by $J_{m,max}$, $\forall m\in\mc M$. Therefore, choosing $\delta_{max}\ge \max_{m\in\mc M} \{J_{m,max}\}$ guarantees the feasibility of the optimization problem.}. For now, we assume that the cost $J_m(x)$ can be computed by each robot $i$. In Section \ref{sec:task_all}, we will discuss how certain assumptions on the structure of the costs $J_m$ can lead to a decentralized solution to \eqref{eq:udeltaqp}. \par
The optimization problem presented in \eqref{eq:udeltaqp}, is solved at each point in time to generate a control input $u_i$ for robot $i$. The following proposition establishes the convergence of the sequence of solutions of the above optimization problem for all the robots.
\begin{proposition}
	Consider a team of $N$ robots that execute $M$ different tasks, robot $i$ implementing the control input $u_i^{(k)}$ at time $k$, obtained by solving the optimization problem \eqref{eq:udeltaqp}, where, besides being Lipschitz, $\gamma$ is assumed to be continuously differentiable. The sequences of solutions of \eqref{eq:udeltaqp}, $\{u_i^{(k)}\}_{k\in\mathbb N},\{\delta_i^{(k)}\}_{k\in\mathbb N}, i \in \mathcal{N}$, converge as $k\to\infty$. Specifically, $u_i^{(k)},\forall i \in \mathcal{N}$ converges to zero, and $\delta_i, \forall i \in \mathcal{N}$ converges to the value $\gamma(J(x))$.
\end{proposition}
\begin{proof}
	Invoking the relation $h_m = -J_m$, the constraint corresponding to each robot in \eqref{eq:udeltaqp} at time $k$ can be rewritten in vectorized form as, 
	\begin{equation*} \label{eq:vec_cons}
	\frac{\partial J}{\partial x_i}u_i \leq  \delta_i - \gamma(J(x))
	\end{equation*}
	where  $J(x) = [J_1(x),J_2(x),\ldots,J_M(x)]^T$, and $\gamma(J(x))$ is intended component-wise.
	For each robot solving the optimization problem presented in \eqref{eq:udeltaqp}, the Lagrangian can be written as
	\begin{equation*}
	L(u_i,\delta_i,\lambda_i) = \|u_i\|^2 + l\|\delta_i\|^2 + \lambda_i^T\left (\frac{\partial J}{\partial x_i}u_i +\gamma(J(x)) - \delta_i\right ).
	\end{equation*}
	From the KKT conditions \cite{boyd2004convex}, we obtain:
	\begin{equation} \label{eqn:l_sol}
	u_i = -l\frac{\partial J}{\partial x_i}^T \delta_i, \quad\lambda = 2l\delta.
	\end{equation}
	The complementary slackness condition gives us
	\begin{equation*} \label{eqn:comp_slack}
	2l\delta_i^T\left(\frac{\partial J}{\partial x_i}u_i - \delta_i + \gamma(J(x))\right) = 0.
	\end{equation*}
	By considering the dual problem, one can show that the following relation holds:
	\begin{equation} \label{eqn:del_J_rel}
	\delta_i = \left( I + l\frac{\partial J}{\partial x_i}\frac{\partial J}{\partial x_i}^T \right)^{-1}\gamma(J(x)),
	\end{equation}
	where $I$ is an $M\times M$ identity matrix.
	In order to show the convergence of the sequences of solutions of the optimization problem, we proceed by defining a Lyapunov candidate function $V: \mathbb{R}^{Nd} \to \mathbb{R}$ as
	\begin{equation*}
	V(x) = \frac{1}{2}\gamma(J(x))\tr \gamma(J(x)).
	\end{equation*}
  Then:
	\begin{equation}
	\label{eq:vdot}
	\dot V =  \gamma(J(x)) \tr \frac{d\gamma}{dJ} \frac{\partial J}{\partial x} \dot x = \gamma(J(x)) \tr \frac{d\gamma}{dJ} \sum_{i\in\mathcal N} \frac{\partial J}{\partial x_i} u_i.
	\end{equation}
	By the definition of $\gamma$, one has that
	\begin{equation}
	\label{eq:gammalipschitz}
	\dfrac{d\gamma}{dJ} \le diag([L,\ldots,L]),
	\end{equation}
	where $L$ is the Lipschitz constant of $\gamma$ and $diag$ is the operator mapping a vector to a diagonal matrix.
	Substituting \eqref{eqn:l_sol} and \eqref{eqn:del_J_rel} into the \eqref{eq:vdot} and using \eqref{eq:gammalipschitz}, we get:
	\begin{equation}
	\label{eq:vdotmess}
	\dot V \le - lL\gamma(J(x))\tr \sum_{i\in\mathcal N}\frac{\partial J}{\partial x_i}\frac{\partial J}{\partial x_i}^T A_i \gamma(J(x)),
	\end{equation}
	where
	\begin{equation*}
	A_i = \left( I + l\frac{\partial J}{\partial x_i}\frac{\partial J}{\partial x_i}\tr \right)^{-1}.
	\end{equation*}
	Owing to the structure of the matrices $A_i$, \eqref{eq:vdotmess} can be rearranged as follows:
	\begin{equation*}
	\dot V \le - lL \sum_{i\in\mathcal N}\gamma(J(x))\tr\frac{\partial J}{\partial x_i} \tilde A_i \frac{\partial J}{\partial x_i}^T \gamma(J(x)),
	\end{equation*}
	where
	\begin{equation*}
	\tilde A_i = \left( \tilde{I} + l\frac{\partial J}{\partial x_i}\tr\frac{\partial J}{\partial x_i} \right)^{-1},
	\end{equation*}
	$\tilde{I}$ being a $d\times d$ identity matrix.
	Because of the symmetry and positive definiteness of the matrices $\tilde A_i$, we can write the upper bound for $\dot V$ in the following way:
	\begin{equation*}
	\dot V = -lL \sum_{i\in\mathcal N}\left\|\frac{\partial J}{\partial x_i}\tr \gamma(J(x)) \right\|_{\tilde A_i}^2 \leq 0,
	\end{equation*}
	where $\|\cdot\|_{\tilde A_i}$ is the norm induced by the inner product $\langle \cdot,\cdot \rangle_{\tilde A_i}$.
	We can now apply LaSalle's invariance principle \cite{khalil2015nonlinear} to show that the sequence of states $x^{(k)}$ of the robots at time $k$ converges to the set 
	\begin{equation*} \label{eqn:inv_set}
	E = \left \{x\in\mathbb R^{Nd} ~\left\vert~ \gamma(J(x)) \in \bigcap\limits_{i\in\mathcal N}~\mathrm{null}\left (\frac{\partial J}{\partial x_i}\tr\right )\right. \right \}.
	\end{equation*}
	Substituting \eqref{eqn:del_J_rel} in \eqref{eqn:l_sol}, we can see that, as $k\to\infty$,
	\begin{equation*}
	x^{(k)}\to\in E \implies u_i^{(k)}\to 0,~\forall i\in\mathcal N.
	\end{equation*}
	Moreover, \eqref{eq:udeltaqp} implies that $\delta_i^{(k)} \to \gamma(J(x^{(k)}))$ as $u_i^{(k)} \to 0$. From \eqref{eqn:l_sol}, it follows that the same results hold for the case when $\delta = 0$.
\end{proof}
 We now present an illustrative example to highlight the effect that the slack variables $\delta_i$ have on the execution of multiple tasks simultaneously. 

\begin{figure}
	\centering
	\subfloat[][]{\label{subfig:threetasksa}\includegraphics[trim={10cm 0cm 10cm 0cm}, clip,width=.2\textwidth]{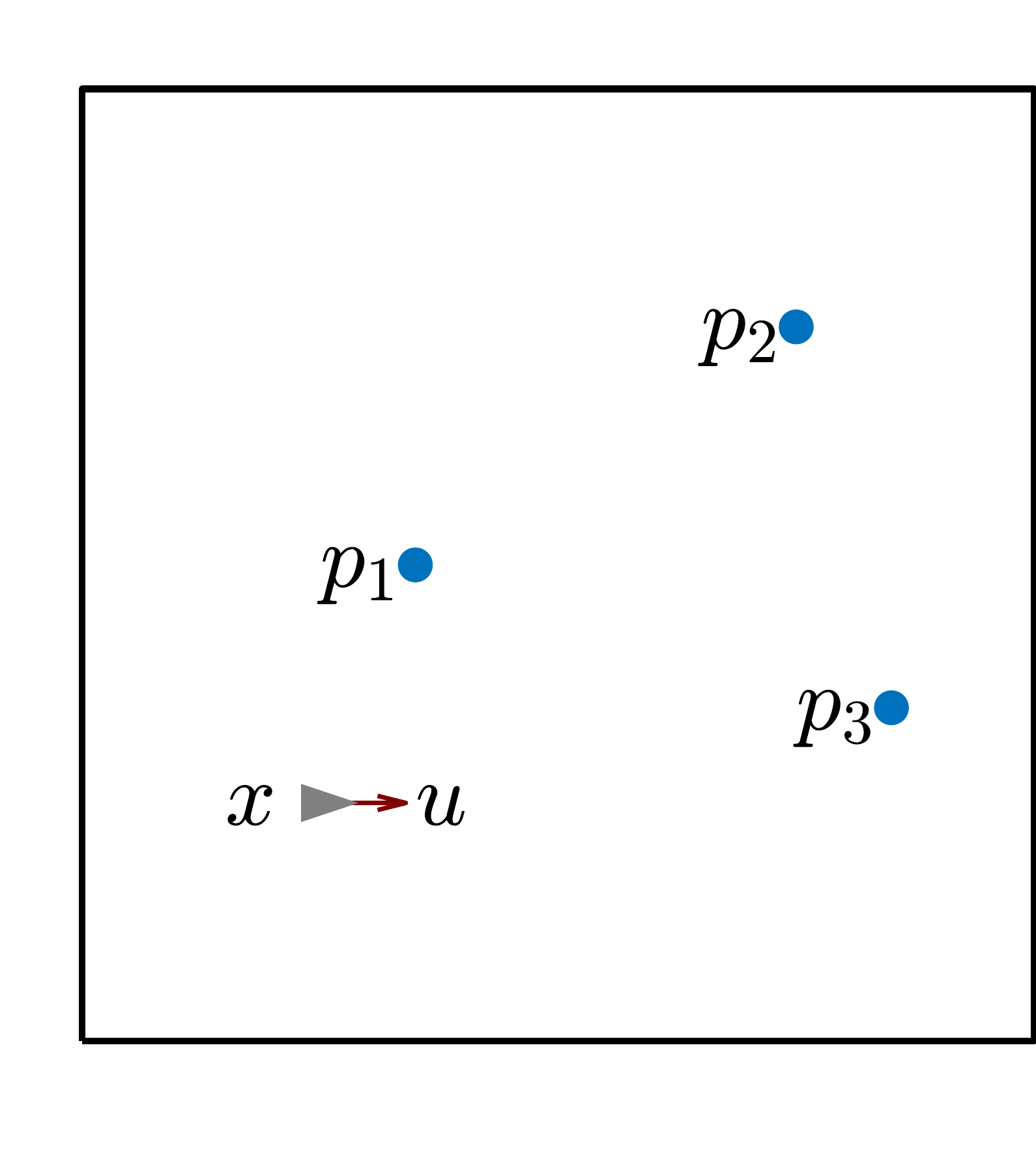}}~
	\subfloat[][]{\label{subfig:threetasksb}\includegraphics[trim={10cm 0cm 10cm 0cm}, clip,width=.2\textwidth]{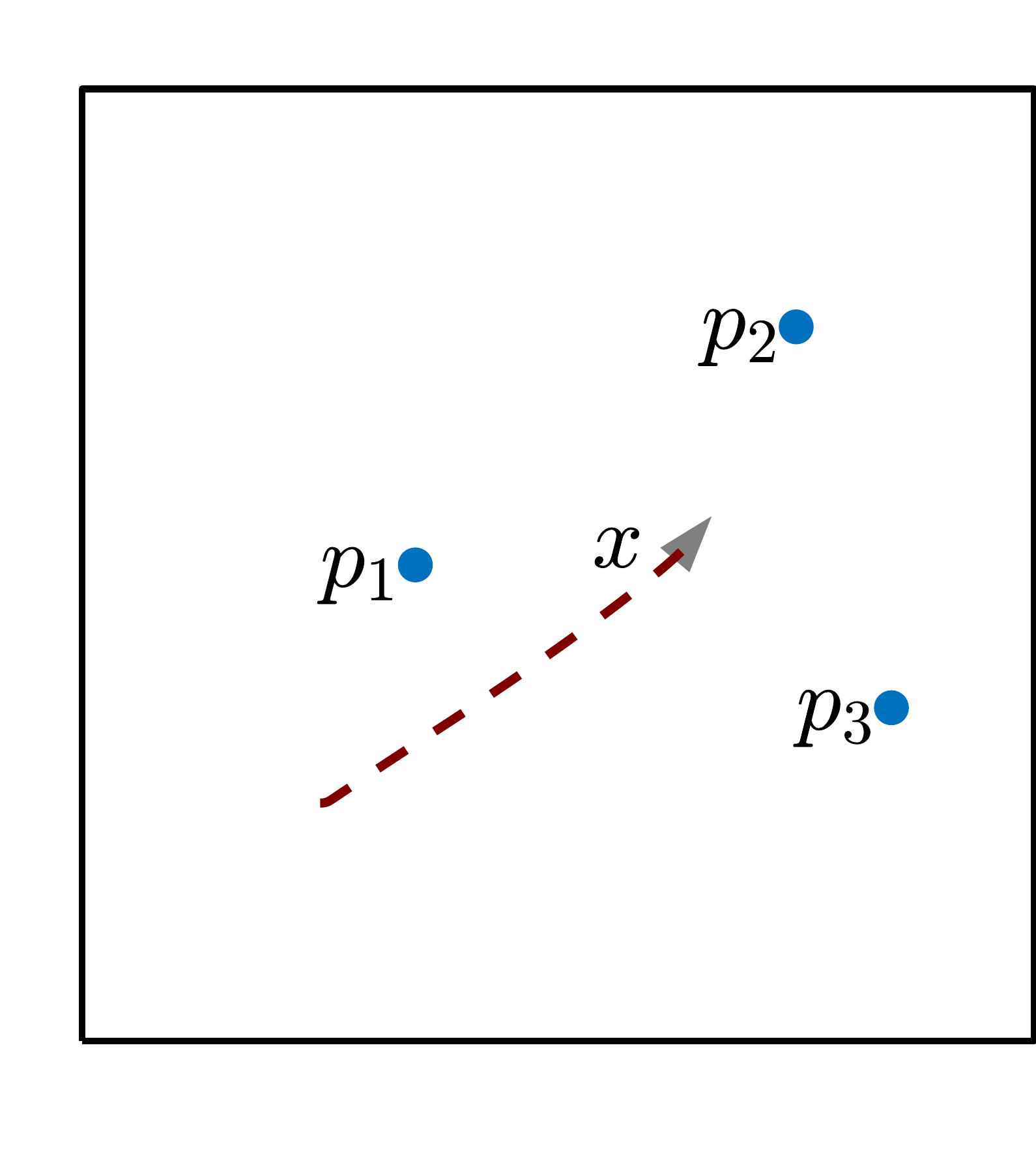}
	\label{subfig_sim1_b}
		}
	\caption{Illustration of Example \ref{exmp:threetasks} for the constraint-based multi-task execution problem formulated in \eqref{eq:udeltaqp}. The robot, represented by the gray triangle is assigned three tasks, which involve driving to the three points $p_1,p_2$ and $p_3$ simultaneously. It is, in fact, impossible to accomplish all three tasks at the same time. We show how, by iteratively solving \eqref{eq:udeltaqp}, the robot drives to a point which is almost equidistant from the three points, as seen in \ref{subfig_sim1_b}. The slack variables $\delta_i$ encode the extent to which the constraints pertaining to each task are violated. Thus, given no preferences on the tasks, the robot chooses to fulfill all three to an equal degree.}
	\label{fig:threetasks}
\end{figure}
\begin{example}\label{exmp:threetasks}
Consider the simplistic case where one robot, moving in a planar environment, has to perform three separate tasks simultaneously, where each task involves driving to a point in the environment. Thus, the robot is tasked with reaching three separate points $p_1$, $p_2$ and $p_3$ simultaneously, depicted as blue dots in Fig.~\ref{fig:threetasks}. The robot is represented by the gray triangle and its position and velocity are denoted by $x$ and $u$, respectively.  The robot implements the optimization program~\eqref{eq:udeltaqp}, with $h_{m}(x) = -\|x-p_m\|^2,~m\in\{1,2,3\}$. Under the effect of the input $u$, solution of \eqref{eq:udeltaqp}, the robot is driven towards a point which is almost equidistant from the three goal points (Fig.~\ref{subfig:threetasksb}).\par 
This example presents an impossible situation -- the robot cannot be present at all three points at the same time. In fact, the purpose is to show how the robot can execute each task, while using the slack variables to relax the constraints for the tasks. 
\end{example}

The components of $\delta_i$ encode the extent to which the corresponding task constraints are relaxed thus signifying the relative effectiveness with which one task is performed over another. We will see later how, by enforcing constraints on the elements of $\delta_i$, we can allow the robots to prioritize between tasks, and ultimately lead to a mechanism to allocate tasks among robots.

\section{Optimal Task Allocation}
\label{sec:task_all}
The constraint-based optimization formulation introduced in the previous section, involved minimizing the control effort expended by a robot, subject to multiple task constraints.  The introduction of slack variables allowed each robot to perform tasks with varying levels of effectiveness. However, in this formulation, robots might end up prioritizing some tasks over others simply as a function of the control effort required for each task. In realistic scenarios, some tasks might be more important than others and might require a higher level of attention from the robots. Consequently, the team might be required to divide itself among the tasks according to a global, system-level specification. \par 
This section develops an optimization framework which allows individual robots to execute multiple tasks while prioritizing some tasks over others. The first formulation does not take into account the heterogeneous capabilities of the robots. Following this, the problem is modified to take advantage of the robot heterogeneity. \par 

As discussed in Section \ref{sec:introduction}, task priorities can be introduced via additional constraints on the slack variables for each task. As an example, for a given robot $i$, performing task $T_m$ with the highest priority would imply that
\begin{equation}\label{eq:deltanm}
\delta_{i,m} \leq \delta_{i,k} \quad\forall k \in \mathcal{M},
\end{equation}
where, as introduced in \eqref{eq:udeltaqp}, $\delta_{i,m}$ represents the extent to which robot $i$ can relax the task constraints corresponding to task $T_m$. We now present an example to illustrate the effect that the priority constraints given by \eqref{eq:deltanm} can have on effectiveness with which a robot performs different tasks.\par 
\begin{figure}
	\centering
	\subfloat[][]{\label{}\includegraphics[trim={10cm 0cm 10cm 0cm}, clip,width=.2\textwidth]{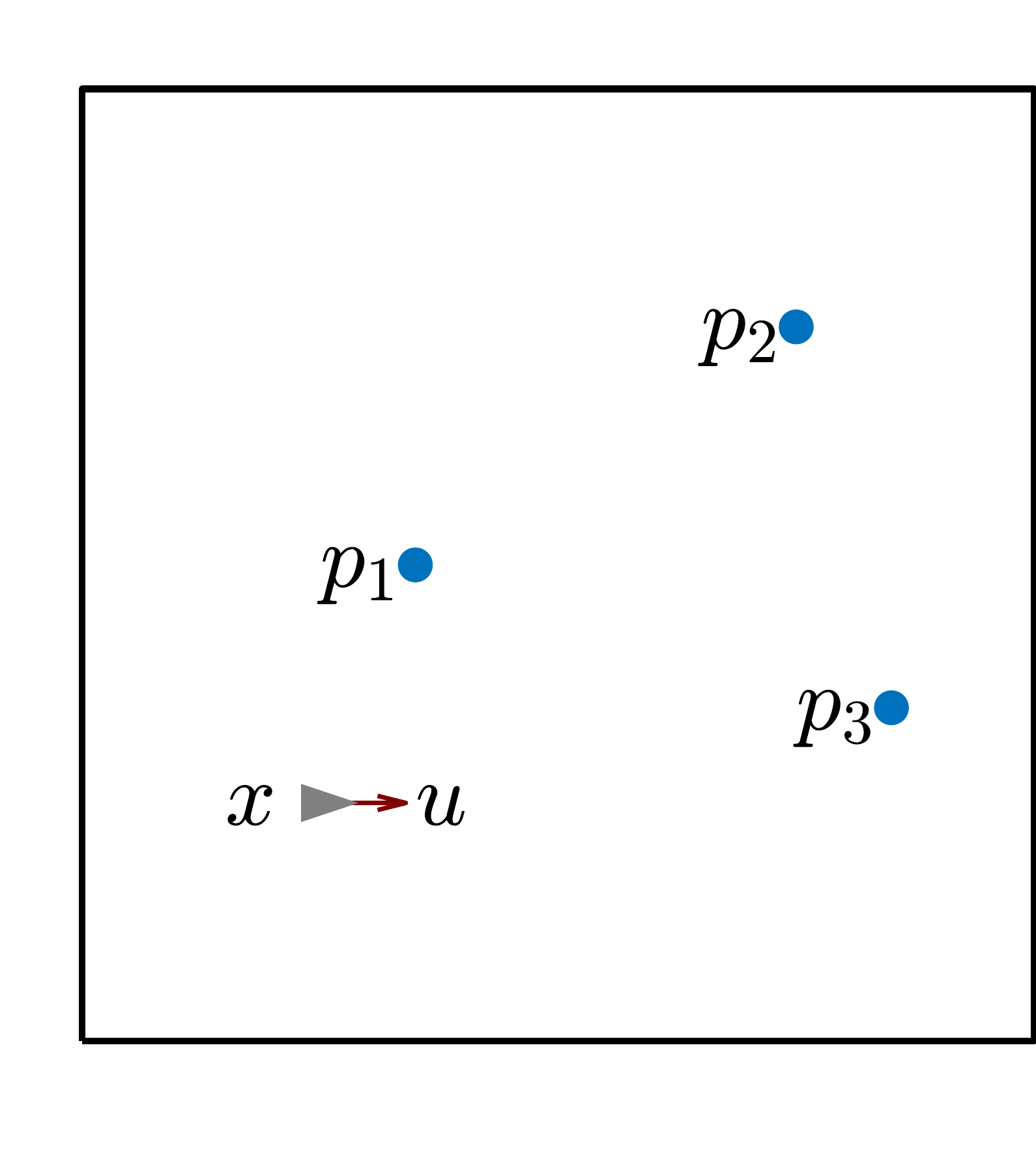}}~
	\subfloat[][]{\label{}\includegraphics[trim={10cm 0cm 10cm 0cm}, clip,width=.2\textwidth]{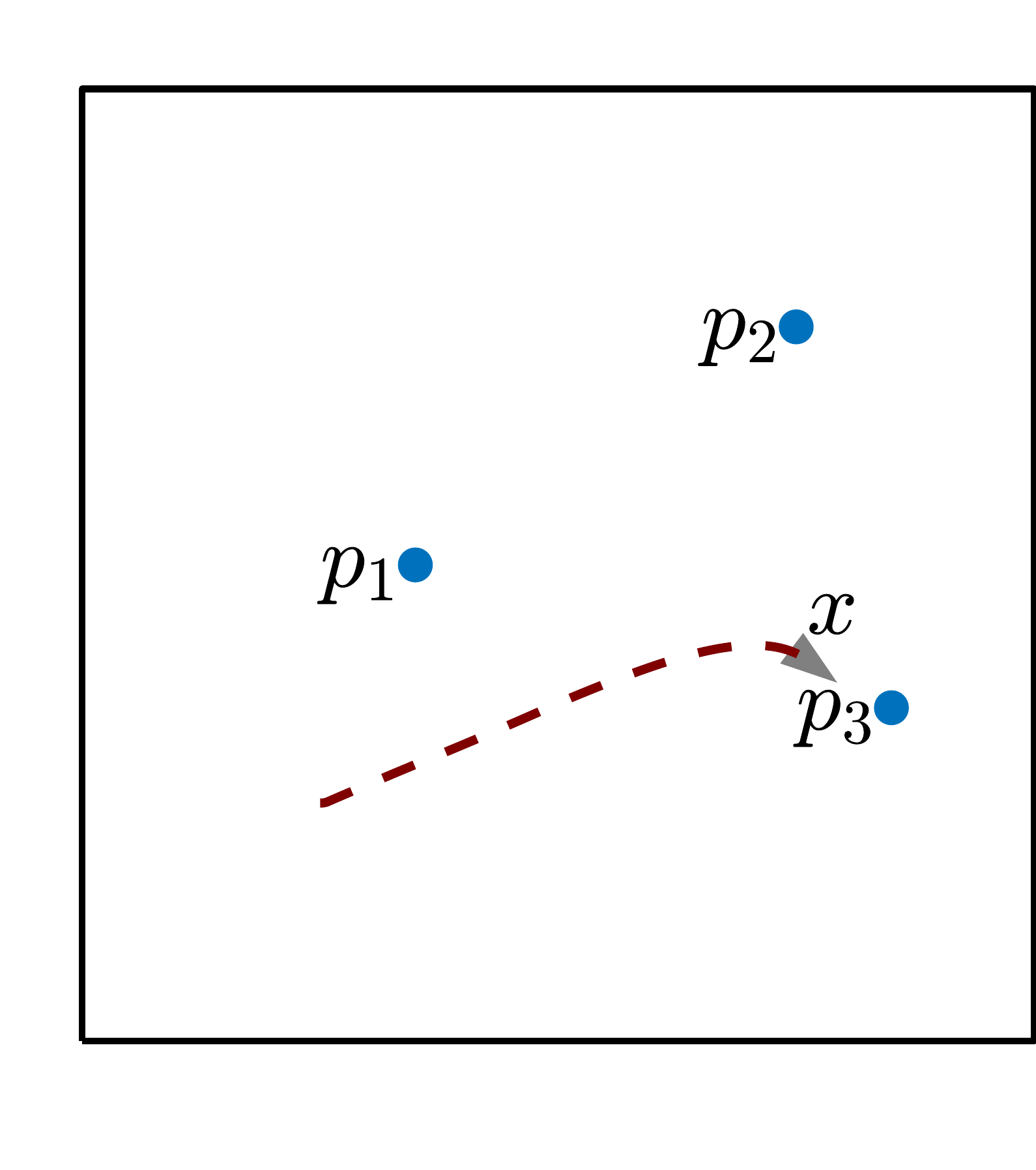}}
	\caption{Illustration of Example \ref{exp_threetasks_const} in order to highlight the effect that additional constraints on the components of the slack variable $\delta_i$ have on the execution of multiple tasks by the robot. Owing to the constraints imposed by \eqref{eq:constraints}, the robot performs task 3 (which involves driving towards point $p_3$) with higher effectiveness than the other two tasks.}
	\label{fig:threetasksconstraints}
\end{figure}
\begin{example} \label{exp_threetasks_const}
	As in Example~\ref{exmp:threetasks}, a robot is tasked with going to three different points ($p_1$, $p_2$ and $p_3$ in Fig.~\ref{fig:threetasksconstraints}) of the environment in which it is deployed. However, differently from Example~\ref{exmp:threetasks}, we now impose additional constraints on the values of $\delta_m,~m\in\{1,2,3\}$ in order for task $T_3$ to have the highest priority. In particular, the constraints are the following:
	\begin{equation}\label{eq:constraints}
	\delta_{1,3} \leq \frac{1}{10}\delta_{1,1}\qquad\text{and}\qquad\delta_{1,3} \leq \frac{1}{10}\delta_{1,2}.
	\end{equation}
	As seen in Fig.~\ref{fig:threetasksconstraints}, the input $u(t)$, solution of the optimization problem \eqref{eq:udeltaqp} with the additional constraints \eqref{eq:constraints}, drives the robot towards $p_3$.
\end{example}

In order to introduce global task allocation specifications, let $\pi_m^*$ denote the desired fraction of robots that need to perform task $T_m$ with highest priority. Then, $\pi^* = [\pi_1^*,\pi_2^*,\ldots,\pi_M^*]^T$ denotes the global task specification for the team of robots. We would like the robots to achieve a trade-off between maximizing their long term operation in the environment and achieving the desired global task allocation. To this end, let $\alpha_i = [\alpha_{i,1},\ldots,\alpha_{i,M}]^T \in \{0,1\}^M$ denote the vector that indicates the priorities of the tasks for robot $i$ as follows:
\begin{equation*}
\alpha_{i,m} = \begin{cases}
1, \quad\text {if task $T_m$ has the highest priority for robot $i$} \\
0, \quad\text{otherwise}.
\end{cases}
\end{equation*}
By definition, at a given point in time, only one element of $\alpha_i$ can be non-zero. Then, it follows that $\boldsymbol{1}\tr \alpha_i = 1\quad\forall i \in \mc{N}$, where $\mc{N}=\{1,\ldots,N\}$ is the robot index set and $\boldsymbol{1}$ is the $M$-dimensional vector whose components are all equal to 1. Moreover, given the priority constraints in \eqref{eq:deltanm} and the definition of $\alpha_{i,m}$, we would like the following implication to hold,
\begin{equation}\label{eq:alphadelta}
\alpha_{i,m} = 1 \quad\Rightarrow\quad \delta_{i,m} \leq \frac{1}{\kappa}\delta_{i,n}\quad \forall n\in \mathcal{M},~n\neq m,
\end{equation}
where $\kappa > 1$ allows us to encode how the task priorities impact the relative effectiveness with which robots perform different tasks. For example, suppose task $T_m$ has the highest priority for robot $i$, i.\,e. $\alpha_{i,m}=1$. Then, \eqref{eq:alphadelta} implies that, larger the value of $\kappa$, the more effectively task $T_m$ will be executed compared to any other task $T_n$. \par 
Let $\alpha = [\alpha_1\tr,\alpha_2\tr,\dots,\alpha_N\tr]\tr \in \{0,1\}^{NM}$ represent the vector containing the task priorities for the entire multi-robot system. Then, the task prioritization of the team at any given point in time is given by
\begin{equation*}
\pi(\alpha) = \begin{bmatrix}
\pi_1(\alpha)\\\vdots\\\pi_M(\alpha)
\end{bmatrix} = \frac{1}{N}\sum_{i = 1}^{N} \alpha_i =
\frac{1}{N} \begin{bmatrix}
I_M, I_M, \ldots, I_M
\end{bmatrix} \alpha,
\end{equation*}
where $I_M$ is the $M\times M$ identity matrix.
 
We propose the following optimization problem whose solution minimizes the difference between the current task prioritization of the multi-robot system $\pi(\alpha)$, and the desired one $\pi^*$, while at the same time, allowing the robots to minimize the consumed energy (proportional to $\|u_i\|^2$) subject to the task constraints (with slackness encoded by $\delta_i$):
\begin{subequations}
\begin{align}
\min_{u,\delta,\alpha} ~& C\|\pi^* - \pi(\alpha)\|^2 + \sum_{i = 1}^{N} \Big( \|u_i\|^2 + \|\delta_i \|^2 \Big) \label{eq:miqp:a}\\ 
 \text{s.t.} ~~& \frac{\partial h_{i,m}}{\partial x_i} u_i \geq -\gamma(h_{i,m}(x)) - \delta_{i,m} \label{eq:miqp:b}\\
&\delta_{i,n} \geq \kappa\big (\delta_{i,m} - \delta_{max}(1 - \alpha_{i,m}) \big ),~~ n\neq m \label{eq:miqp:c}\\
&\boldsymbol{1}\tr\alpha_i = 1 \label{eq:miqp:d}\\ 
&\|\delta_i\|_\infty \leq \delta_{max} \label{eq:miqp:e}\\
&\alpha \in \{0,1\}^{NM} \label{eq:miqp:f}\\
&\hspace{3cm}\forall i \in \mc N,~\forall n,m \in \mc M.
\end{align}
\noeqref{eq:miqp:a}\noeqref{eq:miqp:b}\noeqref{eq:miqp:c}\noeqref{eq:miqp:d}\noeqref{eq:miqp:e}\noeqref{eq:miqp:f}
\end{subequations}
In \eqref{eq:miqp:a}, $C$ is a scaling constant allowing for a trade-off between meeting the global specifications and allowing individual robots to expend the least amount of energy possible. The constraint \eqref{eq:miqp:c} encodes the relation described in \eqref{eq:alphadelta}. We now present an extension of this problem formulation which incorporates the possibility that robots can have heterogeneous task capabilities.

\subsection{Task Allocation in Heterogeneous Robot Teams}
As argued in Section \ref{sec:introduction}, a task allocation algorithm should consider the different capabilities of the robots when assigning task priorities to the robots. We now consider a scenario where different robots have varying suitabilities for different tasks, which we aim to encode into the optimization formulation presented in \eqref{eq:miqp:a}-\eqref{eq:miqp:f}. \par 

To this end, let $s_{i,m}\ge 0$ denote a \textit{specialization} parameter corresponding to the suitability of robot $i$ for executing task $T_m$. In other words, $s_{i,m}>s_{i,n}$ implies that robot $i$ is better suited to execute task $T_m$ over task $T_n$ (for example, this might be because robot $i$ is equipped with specialized sensors to perform task $T_m$ or has a higher battery level to match the requirements of the task). The \textit{specialization} matrix $S_i$ can be then defined as follows:
\begin{equation} \label{eqn_spec_mat}
S_i = diag([s_{i,1},\ldots,s_{i,M}]).
\end{equation}
Since $S_i$ is a diagonal matrix whose entries are all non-negative, we can define the seminorm $\|\cdot\|_{S_i}$ by setting $\|x\|_{S_i}^2 = x\tr S_i x,~x\in\R^M$. Measuring the length of a vector $x$ using the seminorm $\|\cdot\|_{S_i}$ corresponds to weighting each component of $x$ differently and proportionally to the corresponding entry in the matrix $S_i$. Therefore, a natural extension of the optimization problem \eqref{eq:miqp:a}-\eqref{eq:miqp:f} to account for the heterogeneous capabilities of the robots is given by:
\begin{equation}
\label{eq:miqph}
\begin{aligned}
\min_{u,\delta,\alpha} ~& C\|\pi^* - \pi_h(\alpha)\|^2 + \sum_{i = 1}^{N} \Big( \|u_i\|^2 + \|\delta_i \|_{S_i}^2 \Big) \\ 
\text{s.t.} ~~& \eqref{eq:miqp:b}~\text{to}~\eqref{eq:miqp:f}\\
&\hspace{3cm}\forall i \in \mc N,~\forall n,m \in \mc M.
\end{aligned}
\end{equation}
where
\begin{equation*}
\pi_h(\alpha) = \frac{1}{N} \begin{bmatrix}
P_1, P_2, \ldots, P_N
\end{bmatrix} \alpha
\end{equation*}
is the task prioritization of the multi-robot team and explicitly discounts the effect of robots which might prioritize a task $T_m$ without any suitability for it (indicated by a 0 entry in the specialization matrix). This is achieved by projecting the vector of priorities $\alpha_i$ of robot $i$ in the column space of the corresponding specialization matrix $S_i$, through the projector $P_i = S_i S_i^\dagger$, where $S_i^\dagger$ is the Moore-Penrose inverse of $S_i$. This way, if robot $i$ has no suitability to perform task $T_m$, its corresponding projection matrix $P_i$ will be:
\begin{equation*}
P_i = diag([\underbrace{1,\ldots,1}_{m-1},0,\underbrace{1,\ldots,1}_{M-m}]).
\end{equation*}
Consequently, robot $i$ will not be counted in the evaluation of the $m$-th component of the vector $\pi_h(\alpha)$. This would mean that the $m$-th components of the priority vectors $\alpha_j$ of the other robots will have to make up for it in order to minimize the distance from the global task specification vector $\pi^\ast$. In general, the column space of the matrix $\mc P = [P_1,\ldots,P_N]$ indicates the tasks that the multi-robot system has capabilities to execute. If $\mc P$ has full column rank, the multi-robot system can execute all the $M$ tasks.

\subsection{QP Relaxation for Heterogeneous Task Allocation}

As mentioned before, $\alpha$ is a vector of binary variables in the optimization problem \eqref{eq:miqph}. Such a mixed integer quadratic programming (MIQP) significantly increases the complexity of the algorithm. Consequently, this section relaxes the MIQP problem presented in \eqref{eq:miqph} and replaces it with a QP.\par

The constraint \eqref{eq:miqp:c} encodes the relation given in \eqref{eq:alphadelta}, and
illustrates how the components of $\alpha_i$, assuming the integer values 0 or 1, are used to impose constraints on the task priorities for robot $i$. We now propose to relax the integer constraint and let $\alpha\in[0,1]^{NM}$. Thus, a value $\alpha_{i,m} \in (0,1)$ implies a relaxation on the constraint \eqref{eq:alphadelta} whose effect is determined by the value of $\alpha_{i,m}$. The smaller the value of $\alpha_{i,m}$, the less constrained the value of $\delta_{i,m}$. Thus, the current allocation $\pi_h(\alpha)$ can be reinterpreted as the \emph{relative priorities} among the different tasks, with higher components implying that the corresponding task receives a higher priority overall.

With this relaxation, the optimization problem \eqref{eq:miqph} turns into the following QP:
\begin{equation}
\label{eq:qprelax}
\begin{aligned}
\min_{u,\delta,\alpha} ~& C\|\pi^* - \pi_h(\alpha)\|^2 + \sum_{i = 1}^{N} \Big( \|u_i\|^2 + \|\delta_i \|_{S_i}^2 \Big) \\
\text{s.t.} ~~& \eqref{eq:miqp:b}~\text{to}~\eqref{eq:miqp:e}\\
&\alpha \in [0,1]^{NM}\\
&\hspace{3cm}\forall i \in \mc N,~\forall n,m \in \mc M.
\end{aligned}
\end{equation}
This optimization program is executed at each time instant to calculate the control inputs $u_i$ of the robots. The size of the QP in terms of number of optimization variables and constraints grows as $NM^2$ ($N$ being the number of robots and $M$ the number of constraints), therefore it can be solved very efficiently using standard computational techniques \cite{boyd2004convex}. \par
The following proposition provides guarantees on the execution of tasks for a multi-robot system solving the optimization problem presented in \eqref{eq:qprelax}.

\begin{proposition}\label{prop:pointwiseconvergence}\leavevmode
Consider a team of $N$ single-integrator robots performing $M$ different tasks $T_1,\ldots,T_M$. The specialization matrices $S_i, i \in \{1,\ldots,N\}$, as defined in \eqref{eqn_spec_mat}, indicate the abilities of the robots at performing different tasks. Let $\pi^*$ denote the desired allocation of tasks to robots. For a multi-robot system that solves the optimization problem presented in \eqref{eq:qprelax}:
\begin{enumerate}[(a)]
\item At least one task is executed at each point in time, i.e.,
\begin{equation*}
\exists m\in \{1,\ldots,M\} ~\mathrm{s.t.}~\dot{J}_m \le 0
\end{equation*}
\item If all task are accomplished, i.e., 
\begin{equation*}
\frac{\de J_n}{\de x}=0,~\forall n\in \{1,\ldots,M\},
\end{equation*}
 then $u=0$.
\end{enumerate}

\end{proposition}

\begin{proof}
\begin{enumerate}[(a)]
\item At time $t$, consider the constraint \eqref{eq:miqp:b},
\[
-\frac{\de J_n}{\de x} u \ge \gamma(J_n(x)) - \delta_n.
\]
If solving \eqref{eq:qprelax} results in $\gamma(J_n(x)) - \delta_n < 0~\forall n\in\mc M$, then $u$ is not constrained. As a result, $u=0$ and therefore, given the single-integrator dynamics $\dot x = u$, $\dot J_n=\frac{\de J_n}{\de x} u=0~\forall n\in\mc M$. On the other hand, if $\exists m\in\mc M~\mathrm{s.t.}~\gamma(J_m(x)) - \delta_m \ge 0$, then
\[
\dot J_m = \frac{\de J_m}{\de x}u \le -(\gamma(J_m(x)) - \delta_m) \le 0,
\]
i.\,e. task $T_m$ is executed. Therefore, at each moment in time, there exists a task that is being executed.
\item From the hypothesis, it follows that the inequality
\[
-\frac{\de J_n}{\de x} u \ge \gamma(J_n(x)) - \delta_n
\]
does not constrain $u$. Thus, solving \eqref{eq:qprelax} yields $u=0$.
\end{enumerate}
\end{proof}

\begin{figure}
\centering
\subfloat[][]{\label{}\includegraphics[trim={10cm 0cm 10cm 0cm}, clip,width=.2\textwidth]{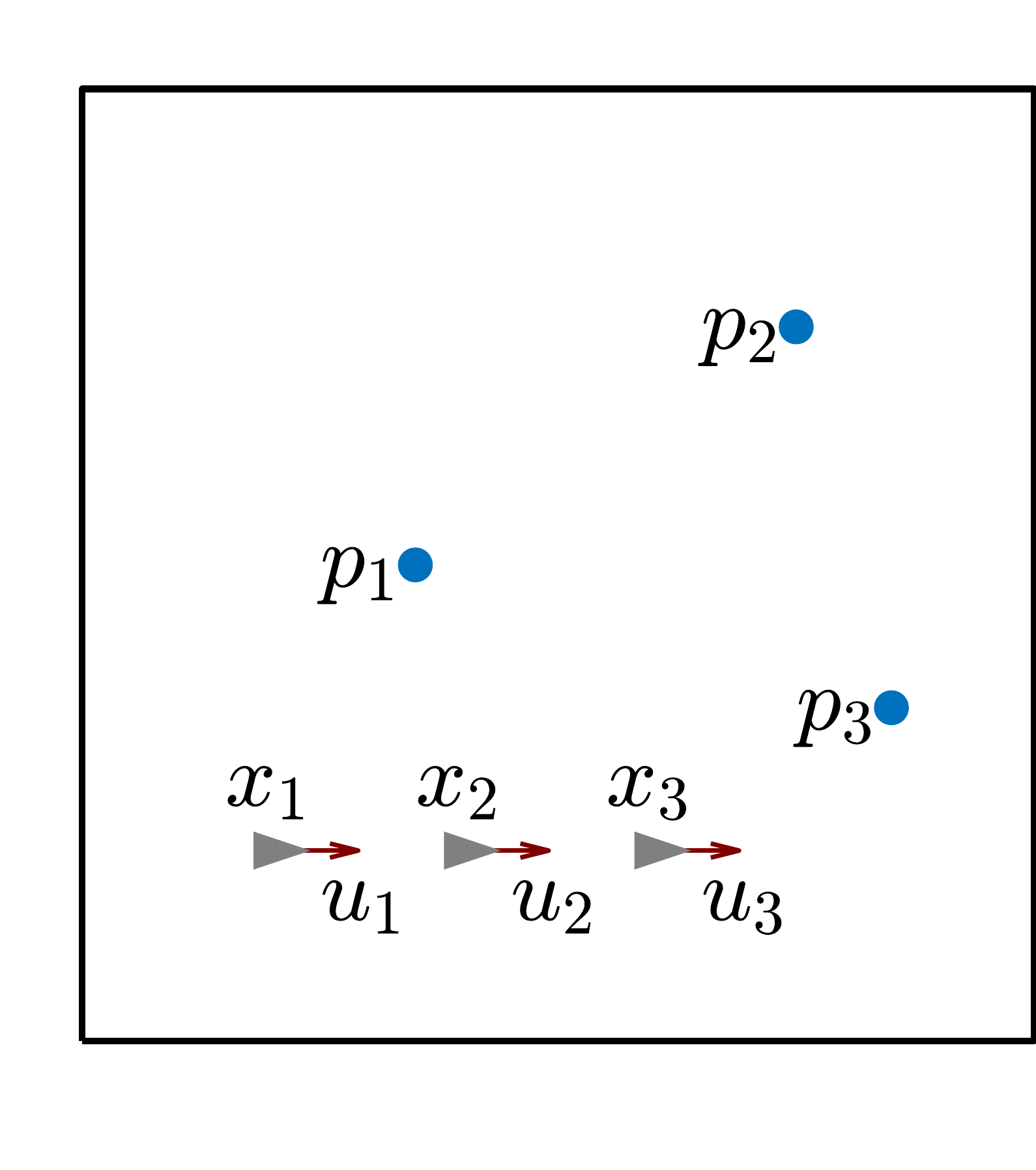}
	\label{subfig_a}
	}~
\subfloat[][]{\label{}\includegraphics[trim={10cm 0cm 10cm 0cm}, clip,width=.2\textwidth]{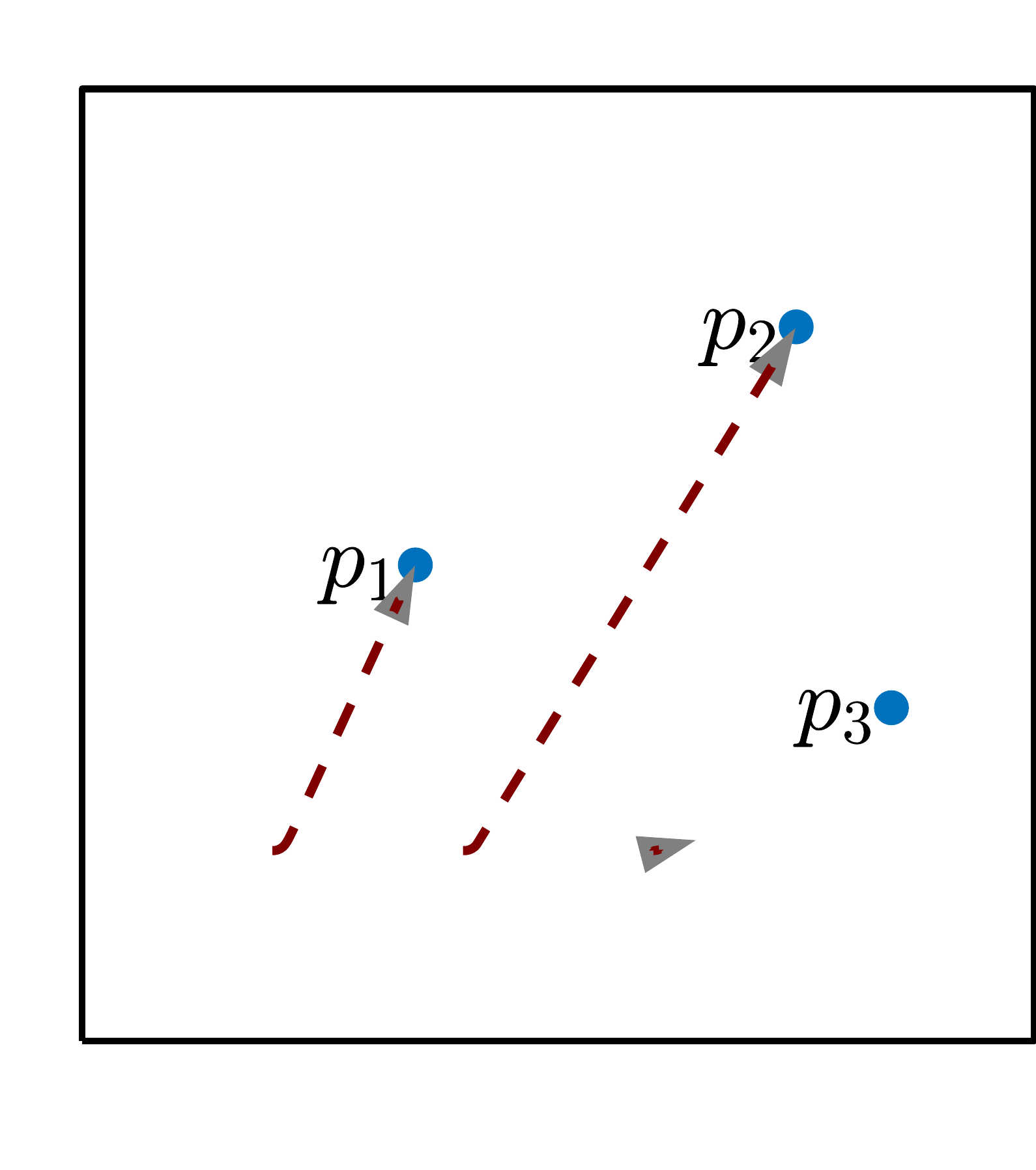}
\label{subfig_b}
	}
\caption{Illustration of the developed task allocation algorithm for a team of three robots tasked to perform three tasks described in Example \ref{exp_hetero}. The robots have heterogeneous capabilities, as encoded by the specialization matrices in \eqref{eqn_s_ex}. In particular, each robot $i$ is specialized towards tracking only one point $p_i$. For a desired task specification $\pi^*=[0.5,0.5,0]\tr$, the trajectories of the robots are depicted in \ref{subfig_b}. As seen, since there is no requirement to perform the third task, robot $3$ moves very little when compared to the other two robots.}
\label{fig:threetasksthreerobots}
\end{figure}

\begin{figure}
	\centering
	\subfloat[][]{\label{subfig:threethreea}\includegraphics[trim={0cm 0cm 0cm 0cm}, clip,width=.24\textwidth]{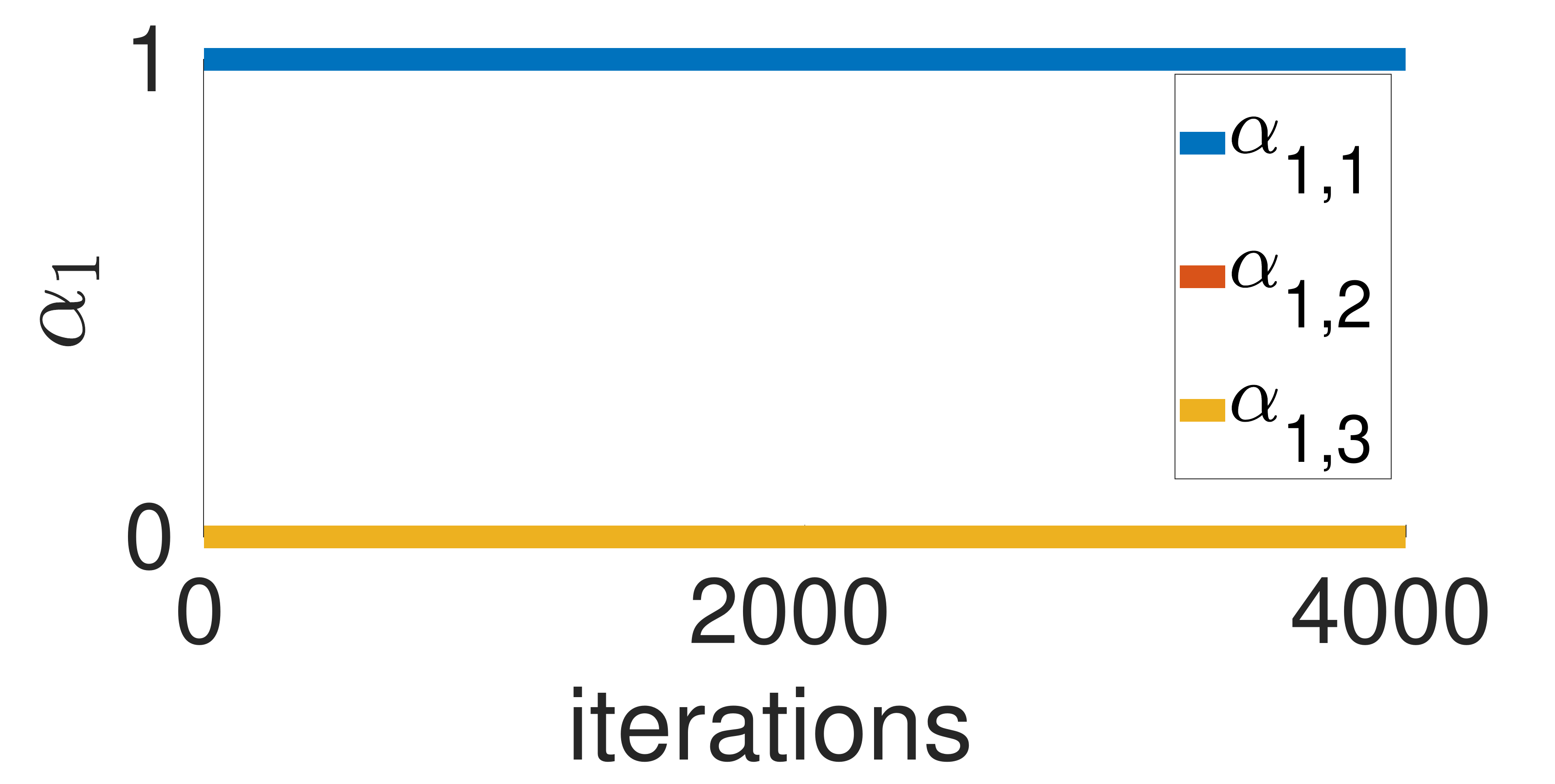}
	\label{sufig_alpha}
		}~
	\subfloat[][]{\label{subfig:threethreeb}\includegraphics[trim={0cm 0cm 0cm 0cm}, clip,width=.24\textwidth]{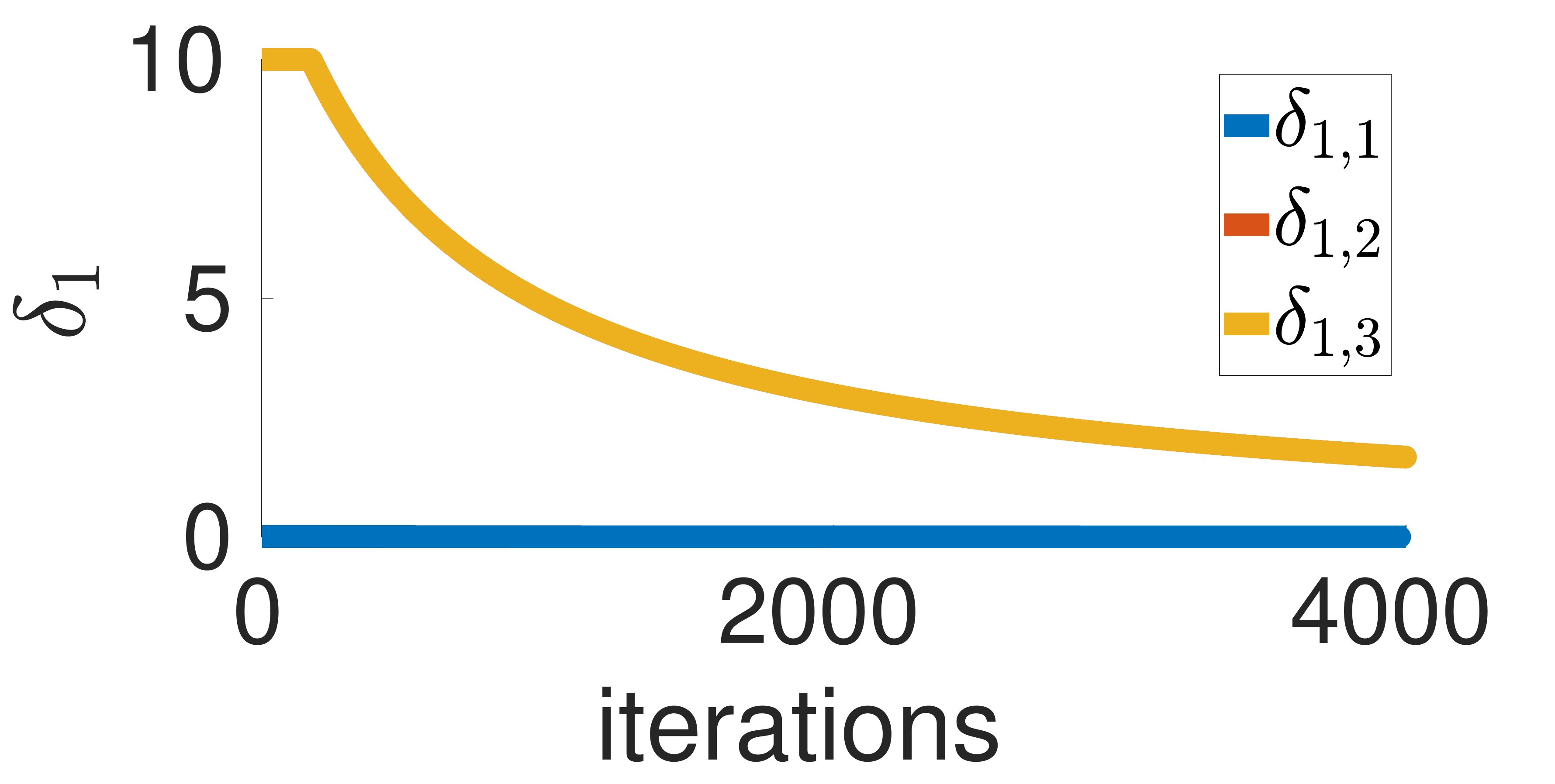}
	\label{subfig_delta}
		}
	\caption{Evolution of the slack variable $\delta_1$ and the relaxed task priorities $\alpha_1$ with time for the task allocation scenario presented in Example \ref{exp_hetero}. As specified by \eqref{eqn_s_ex}, robot $1$ is specialized towards performing task $1$ and consequently, it chooses to perform task $T_1$ with the highest priority, as depicted by the value $\alpha_1 = 1$ in \ref{sufig_alpha}. Accordingly, the slack variable for robot $1$ corresponding to task $1$, denoted as $\delta_{1,1}$ is identically zero, as seen in \ref{subfig_delta}.}
	\label{fig:alphadelta1}
\end{figure}

\begin{example} \label{exp_hetero}
As a final example before showing the experimental results in Section~\ref{sec:exp}, we present the case of three robots that have to execute three different tasks (see Fig.~\ref{fig:threetasksthreerobots}). Each robot has a strong specialization towards executing only one task, in particular robot $i$ is highly specialized to execute $T_i$. The specialization matrices are given by
\begin{align}  \label{eqn_s_ex}
\hspace{-0.7em}S_1=\begin{bmatrix}
1&0&0\\
0&0&0\\
0&0&0
\end{bmatrix} S_2=\begin{bmatrix}
0&0&0\\
0&1&0\\
0&0&0
\end{bmatrix} S_3=\begin{bmatrix}
0&0&0\\
0&0&0\\
0&0&1
\end{bmatrix}.
\end{align}
Task $T_i$ consists in driving to point $p_i$. The global specification vector is set to $\pi^*=[0.5,0.5,0]\tr$, namely there is equal need of executing tasks $T_1$ and $T_2$ and no need to execute task $T_3$ at all. By running the optimization problem \eqref{eq:qprelax}, the robots move as shown by the red trajectories in Fig.~\ref{subfig:threethreeb}. Robot 1 and 2, which can contribute to match the global specification $\pi^*$, move towards points 1 and 2, respectively. Robot 3, whose ability of executing task 3 will not be reflected in any decrease of the first term of the cost in \eqref{eq:qprelax}, moves very little when compared to the other robots.

Fig.~\ref{fig:alphadelta1} shows the values of $\alpha_1$ and $\delta_1$ (related to robot 1). As can be seen, $\alpha_{1,1}$ is identically equal to 1, whereas $\alpha_{1,2}$ and $\alpha_{1,3}$ are identically zero indicating that the robot, owing to its specialization towards task $T_1$, enforces constraints to perform task $T_1$ with a priority higher than $T_2$ and $T_3$. For robot 1, the corresponding values of $\delta_{1,1}$, $\delta_{1,2}$ and $\delta_{1,3}$ are reported in Fig.~\ref{subfig:threethreeb} ($\delta_{1,2}\approx\delta_{1,3}$).
\end{example}

\begin{observation}[Distributed Implementation of Task Allocation Algorithm]
In order to solve the optimization problem \eqref{eq:qprelax} in a distributed fashion, two conditions have to be met:
\begin{enumerate}[(i)]
	\item the constraints \eqref{eq:miqp:b} encoding the tasks have to be decentralized
	\item each robot has to be able to estimate the current task distribution $\pi_h(\alpha)$.
\end{enumerate}
With regards to condition (i), assume that the costs $J_m$ have the following structure:
\begin{equation*}
J_m(x) = \sum_{i = 1}^{N}\sum_{j \in \mathcal{N}_i} J_{i,m}(\|x_i - x_j\|),
\end{equation*}
i.e. they can be broken down into the sum of pairwise costs among each robot $i$ and its neighboring robots $j \in \mathcal{N}_i$. In \cite{cortes2017coordinated}, the authors illustrate the various multi-robot tasks represented by such a cost, and demonstrate the decentralized nature of the control law obtained by performing a gradient-descent on this cost. Furthermore, in \cite{arxiv:extendedacc}, we showed that in this case, the constraint-based task execution can be rendered decentralized provided that certain conditions on the extended class $\mathcal{K}$ function gamma that appears in \eqref{eq:miqp:b} are satisfied.

As far as condition (ii) stated above is concerned, every robot has to have access to $\pi_h(\alpha)$, which encodes the allocation of the entire swarm. The value of $\pi_h(\alpha)$ can either be estimated indirectly, or computed using a distributed consensus protocol, e.g. \cite{ren2008distributed}. This process need not run synchronously with the task allocation and, therefore, if fast enough, can allow a distributed implementation of the optimization program \eqref{eq:qprelax}.
\end{observation}

\section{Experiments} \label{sec:exp}
The presented task allocation algorithm has been implemented on a team of real robots on the Robotarium, a remotely accessible swarm robotics test bed \cite{pickem2017robotarium}. The experimental setup consists in 10 differential-drive robots that move in a 2.5m$\times$1.5m rectangular domain and are asked to perform 2 tasks: environment surveillance and formation control. Task $T_1$ is realized by implementing the coverage control algorithm proposed in \cite{cortes2004coverage}, whereas task $T_2$ consists in driving to specified locations in the domain. \par 
More in particular, as discussed in \cite{cortes2004coverage}, the domain in which the robots move is partitioned into $N$ Voronoi cells corresponding to the $N$ robots. Robot $i$ is in charge of surveiling only its Voronoi cell $V_i$. As shown in \cite{cortes2017coordinated}, the cost to minimize in order to execute this surveillance task is given by
\begin{equation*}
J_1(x) = \sum_{i=1}^{N}\frac{1}{2} \|x_i-G_i(x)\|^2,
\end{equation*}
where $G_i$ denotes the centroid of the Voronoi cell $V_i$.
For task $T_2$, similarly to what has been done in Examples~\ref{exmp:threetasks},~\ref{exp_threetasks_const}~and~\ref{exp_hetero}, the assembly of a fixed formation in space can be encoded by the following cost:
\begin{equation*}
J_2(x) = \sum_{i=1}^{N}\frac{1}{2} \|x_i-y_i\|^2,
\end{equation*}
where $y_i,~i=1,\ldots,N$ are $N$ locations in the workspace which constitute the formation. We consider two different scenarios to illustrate the salient features of the developed task allocation algorithm. In both the scenarios, the robots are assumed to be able to evaluate $\pi_h(\alpha)$.

\subsubsection{Case 1--Heterogeneous robots, two tasks} In this scenario, the robots are asked to perform all tasks: this is realized by setting $\pi^*=[0.5,0.5]\tr$. The robots are characterized by three different specialization matrices: $S_i=diag([1,0])$ for $i\in\{1,3,5,7\}$, $S_i=diag([0,1])$ for $i\in\{2,4,6,8\}$, and $S_i=diag([0.5,0.5])$ for $i\in\{9,10\}$.
\begin{figure*}
\centering
\subfloat[][]{\label{subfig:case1:a}\includegraphics[trim={190px 90px 380px 110px}, clip,width=.245\textwidth]{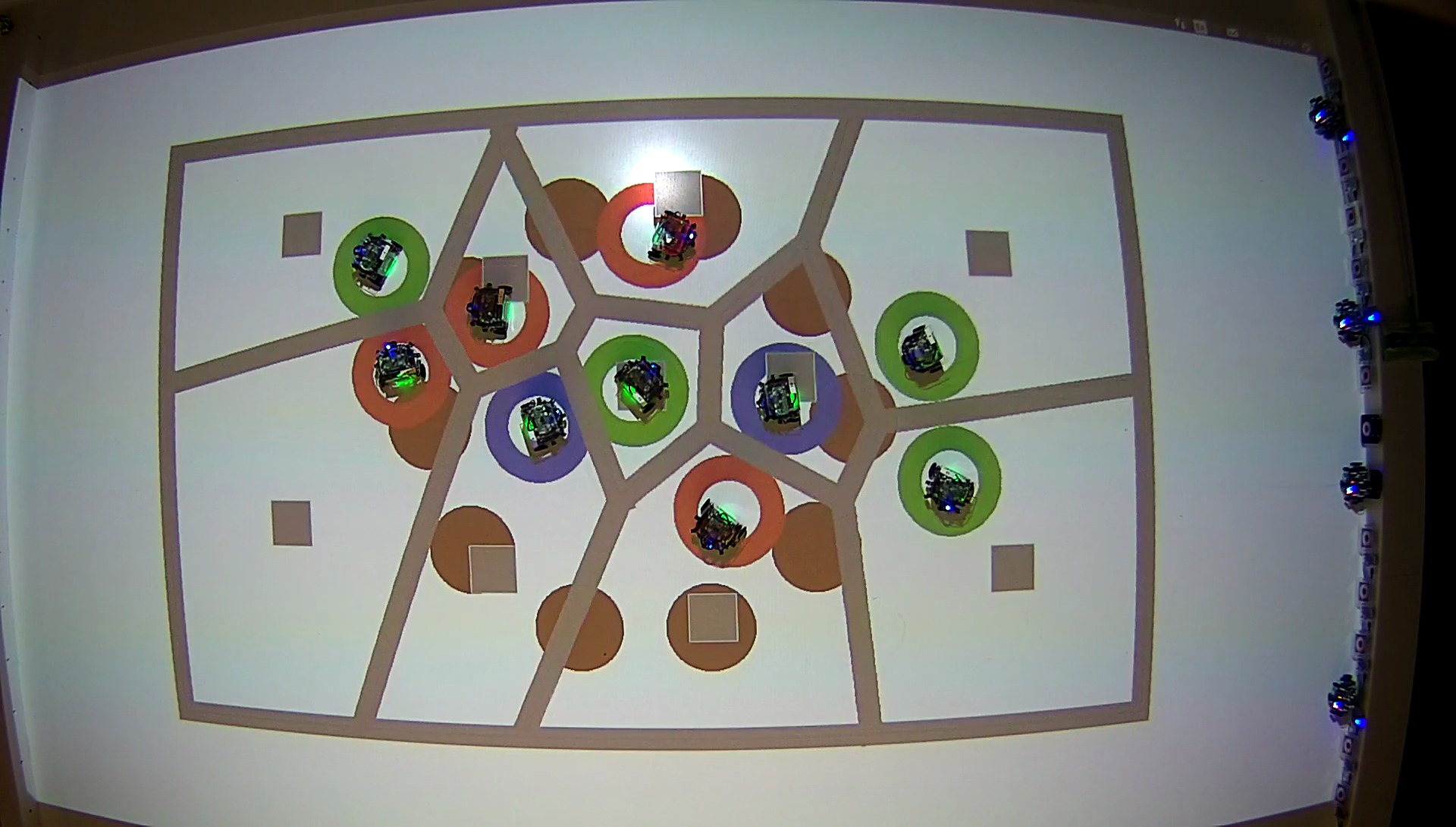}\label{subfig:case2a}}\hfill
\subfloat[][]{\label{subfig:case1:b}\includegraphics[trim={190px 90px 380px 110px}, clip,width=.245\textwidth]{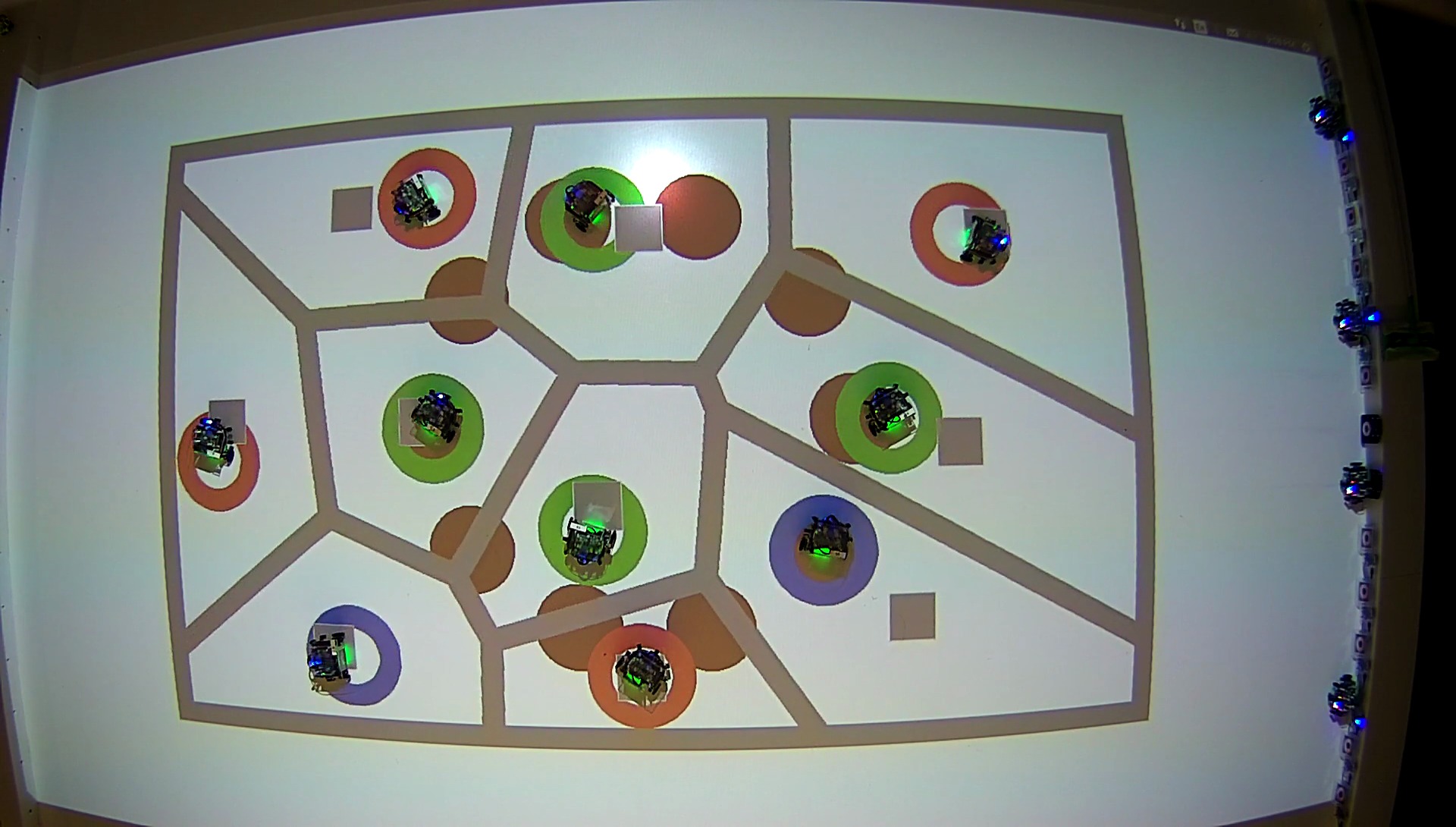}\label{subfig:case2b}}\hfill
\subfloat[][]{\label{subfig:case1:c}\includegraphics[trim={190px 90px 380px 110px}, clip,width=.245\textwidth]{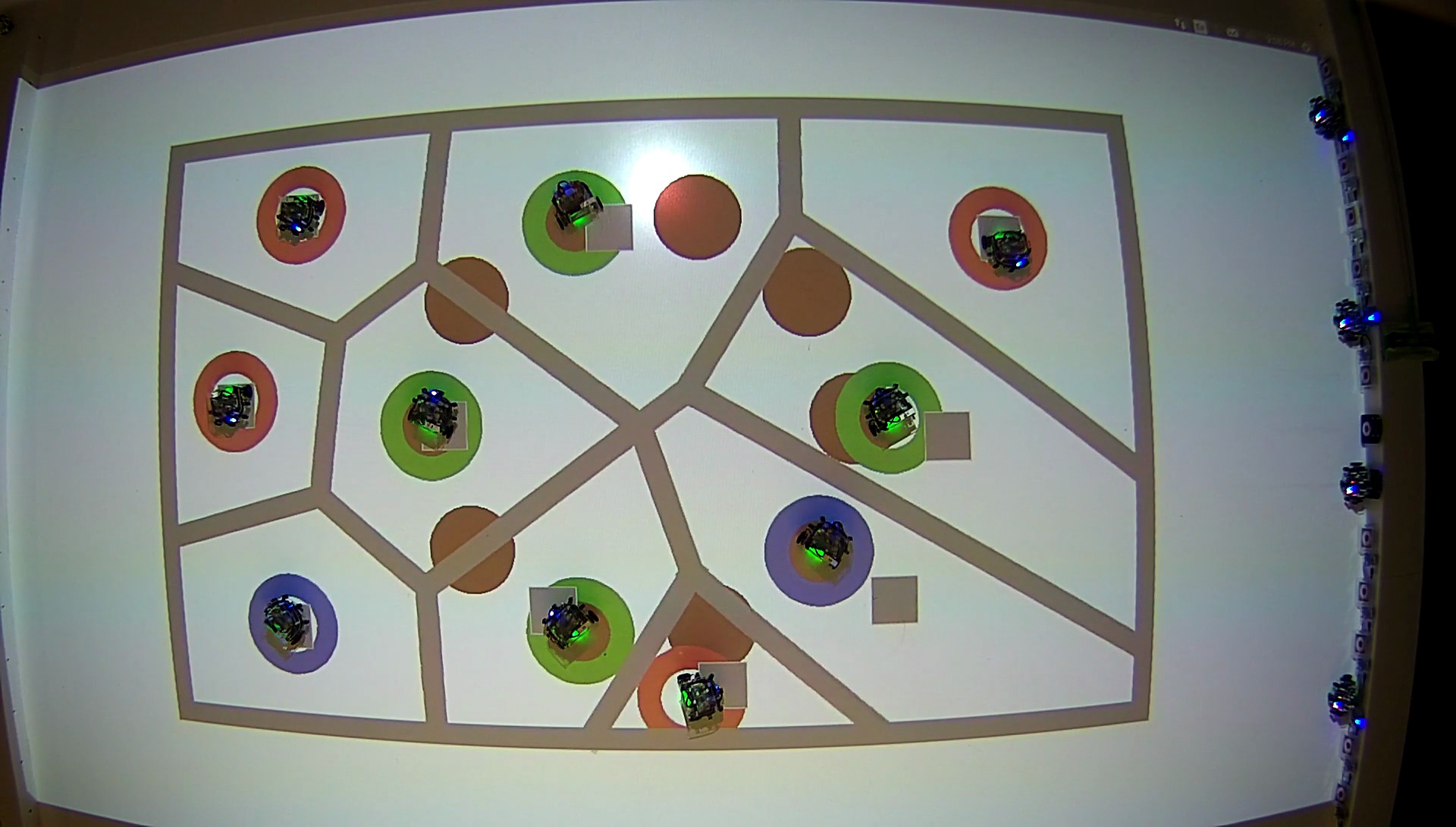}\label{subfig:case2d}}\hfill
\subfloat[][]{\label{subfig:case1:d}\includegraphics[trim={190px 90px 380px 110px}, clip,width=.245\textwidth]{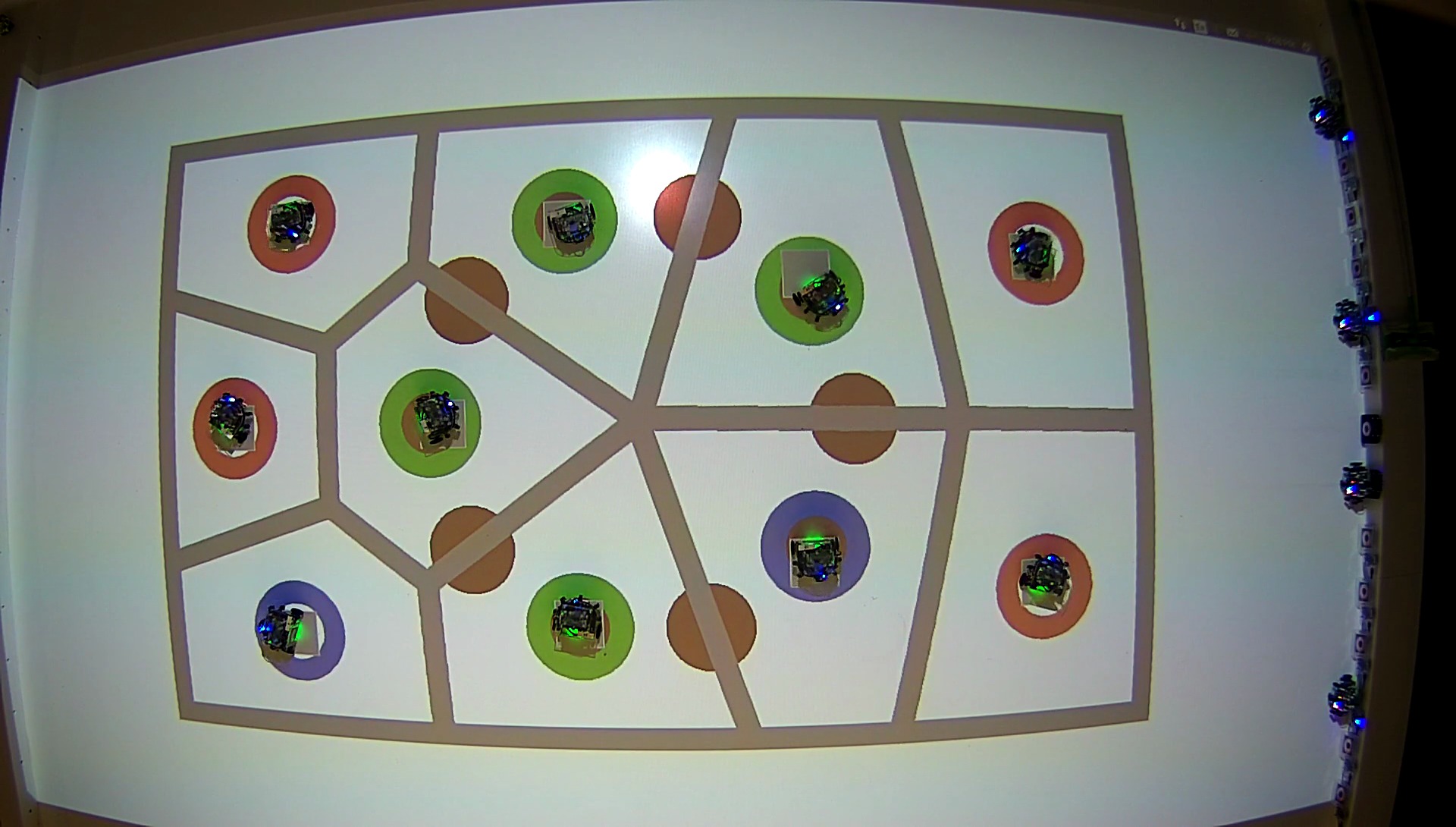}\label{subfig:case2e}}
\caption{Case 1--Heterogeneous robots, two tasks: The snapshots from the experiment performed on the Robotarium show 10 differential drive robots asked to perform two tasks, environment surveillance and formation control. The robots are heterogeneous in their capabilities, with some being specialized to perform surveillance (circled in red in Fig.~\protect\ref{subfig:case1:d}), formation control (circled in green in Fig.~\protect\ref{subfig:case1:d}), and some equally capable of performing both tasks (circled in blue in Fig.~\protect\ref{subfig:case1:d}). The control input $u$, calculated by solving \eqref{eq:qprelax}, drives the robots to the configuration in Fig.~\protect\ref{subfig:case1:d}: here the robots circled in red are close to the centroids of their Voronoi cells (depicted as gray squares), whereas the robots circled in green reached the locations corresponding to a circle formation and indicated by the red circles. The two robots circled in blue did not have bias in their specialization: as a consequence, one of the them is assigned to formation (the rightmost one) and the other one to surveillance. A video of the experiments can be found at \texttt{https://youtu.be/OQiLbEaZsZw}.}
\label{fig:case1}
\end{figure*}
Fig.~\ref{fig:case1} shows snapshots from the video of the experiment performed on the Robotarium. The thick gray lines indicate the boundaries of the Voronoi cells corresponding to the robots, whereas the centroids $G_i$ are depicted as gray squares. The red circles are the locations $y_i,~i=1,\ldots,10$, of the fixed circle formation. The robots, initialized at random locations in the domain (Fig.~\ref{subfig:case1:a}), execute the control input $u_i$ calculated by solving the QP~\eqref{eq:qprelax}, until they reach the configuration shown in Fig.~\ref{subfig:case1:d}. At this point, the robots with $S_i=diag([1,0])$ (circled in red) are on top of the centroids of their Voronoi cells, while the robots with $S_i=diag([0,1])$ (circled in green) have reached their corresponding $y_i$ in the circle formation. For the robots whose $S_i=diag([0.5,0.5])$ (circled in blue), the task allocation obtained by solving the optimization program \eqref{eq:qprelax} resulted in one of them performing surveillance and the other performing formation control.

\subsubsection{Case 2--Homogeneous robots, task switching} In this scenario, the robots have equal specialization matrices and equal suitabilities for all the tasks, i.\,e. $S_i=I_M~\forall i\in\mc N$ ($I_M$ being the $M\times M$ identity matrix). The global specification vector $\pi^*$ is changed at a given point in time from $\pi^*=[1,0]\tr$ to $\pi^*=[0,1]\tr$.
\begin{figure*}
\centering
\subfloat[][]{\label{subfig:case2:a}\includegraphics[trim={190px 90px 380px 110px}, clip,width=.245\textwidth]{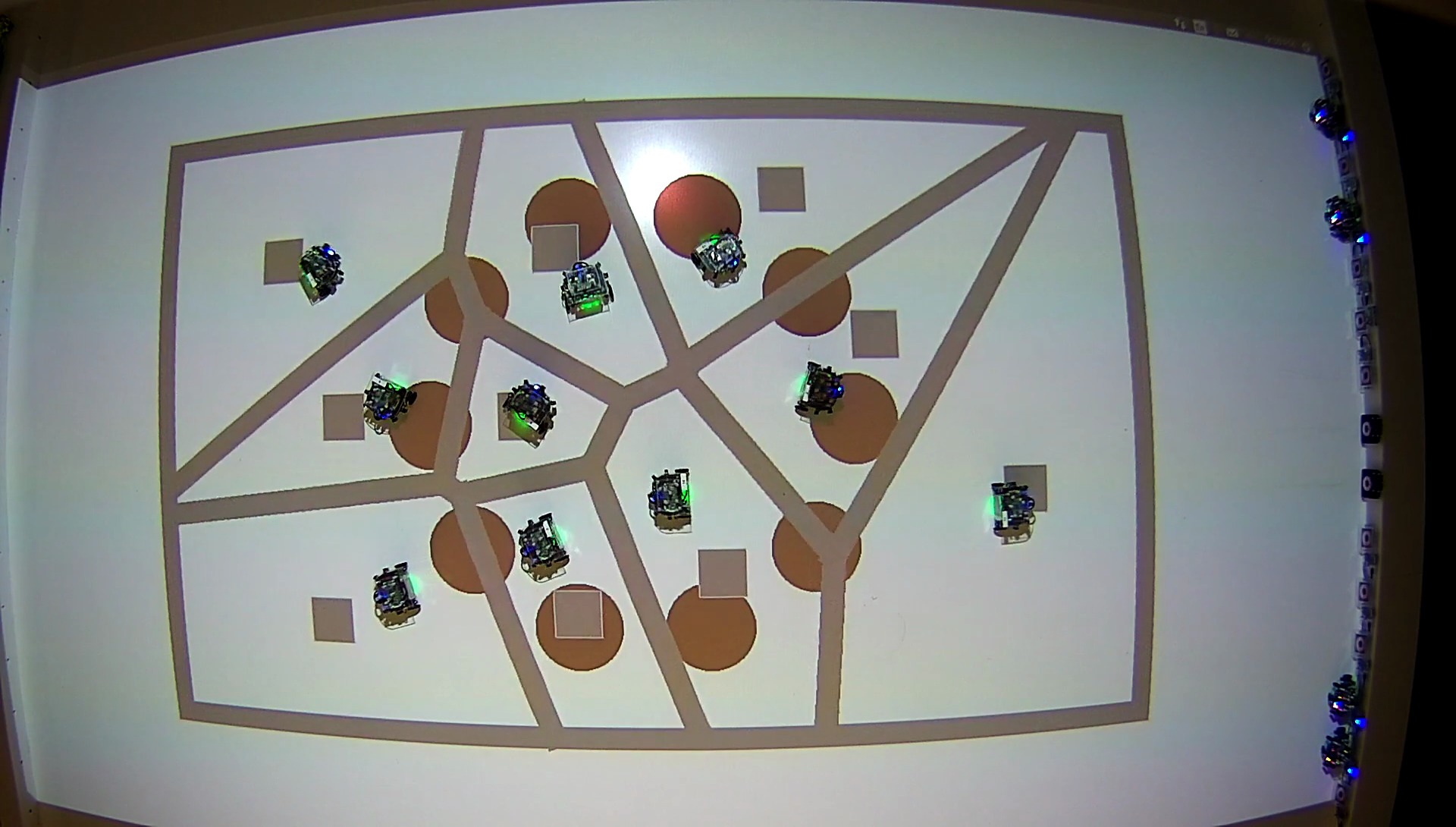}\label{subfig:case4a}}\hfill
\subfloat[][]{\label{subfig:case2:b}\includegraphics[trim={190px 90px 380px 110px}, clip,width=.245\textwidth]{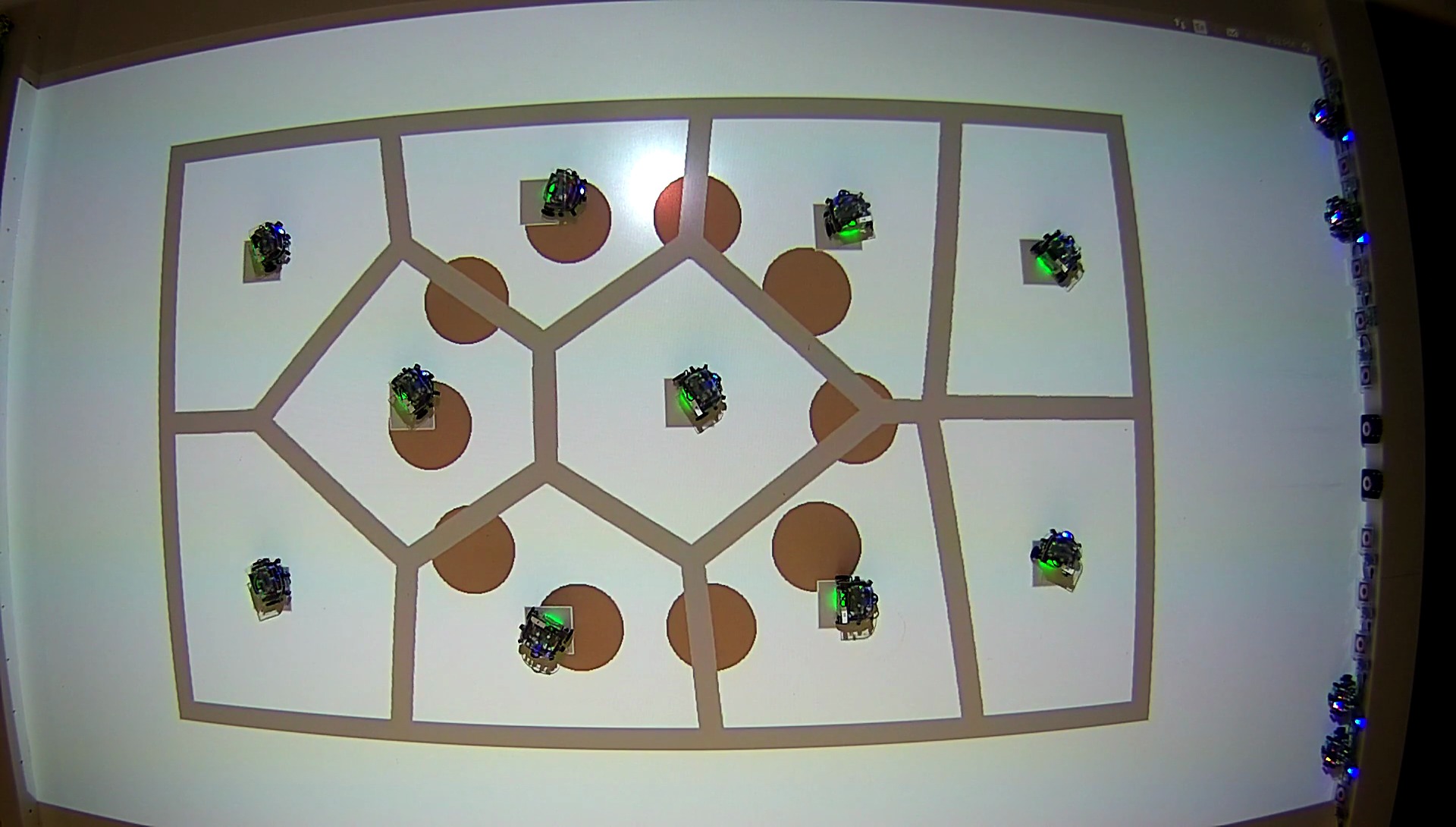}\label{subfig:case4b}}\hfill
\subfloat[][]{\label{subfig:case2:c}\includegraphics[trim={190px 90px 380px 110px}, clip,width=.245\textwidth]{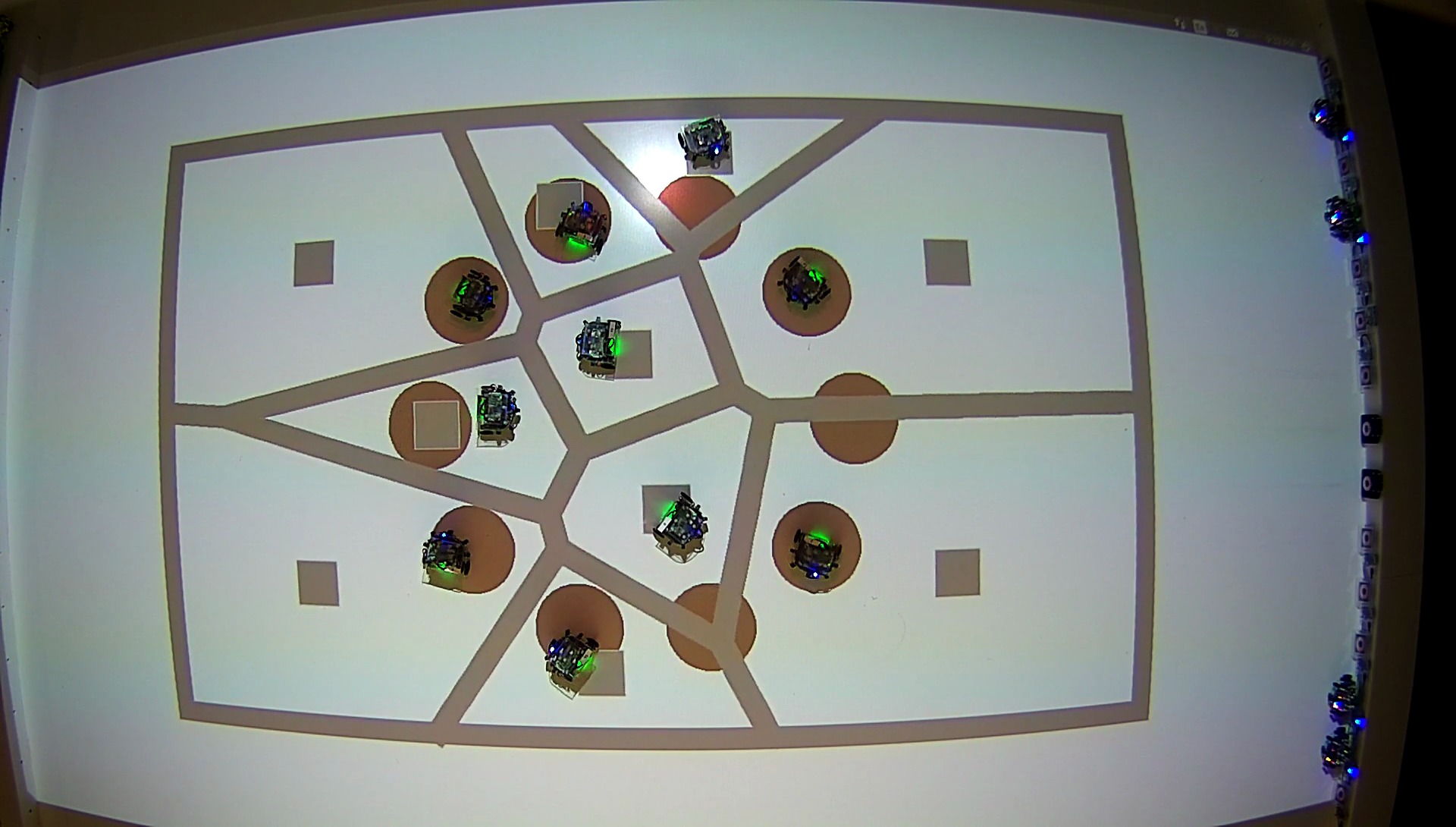}\label{subfig:case4d}}\hfill
\subfloat[][]{\label{subfig:case2:d}\includegraphics[trim={190px 90px 380px 110px}, clip,width=.245\textwidth]{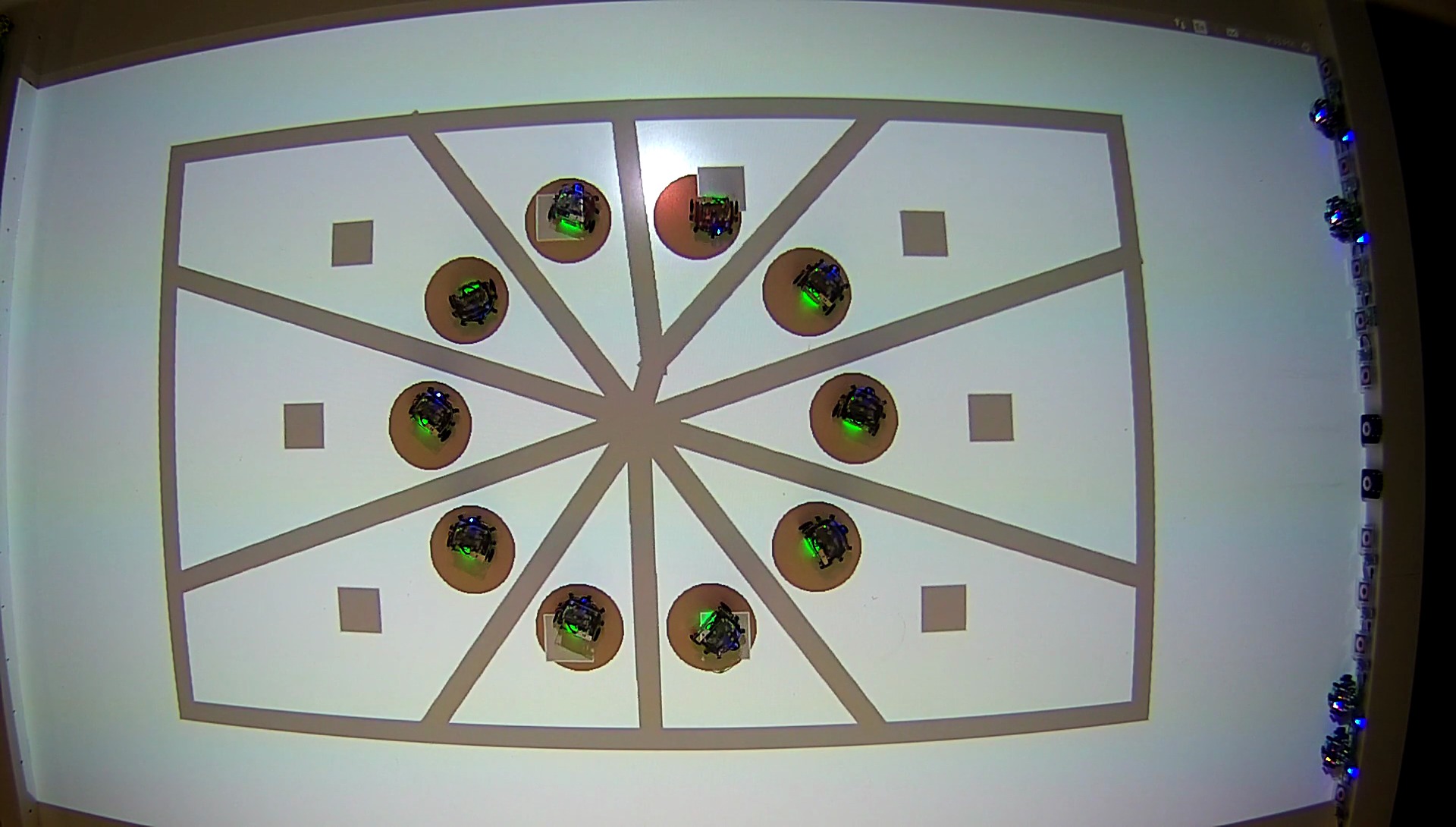}\label{subfig:case4e}}
\caption{Case 2--Homogeneous robots, task switching: In the experiments depicted in these video snapshots, a team of 10 homogeneous robots (characterized by the same specialization matrix $S_i$) is initially asked to perform surveillance of a rectangular domain ($\pi^*=[1,0]\tr$). Under this global task specification, the robots, executing the controller evaluated by solving \eqref{eq:qprelax}, reach a centroidal Voronoi tessellation in Fig.~\protect\ref{subfig:case2:b}. After the first task is accomplished, the global task specification vector is changed to $\pi^*=[0,1]\tr$, corresponding to formation control. Fig.~\protect\ref{subfig:case2:c} shows the robots moving towards the red locations corresponding to a circle formation, which is reached in Fig.~\protect\ref{subfig:case2:d}. This experiment illustrates how the developed optimization framework can dynamically allocate tasks to robots based on changing global specifications. A video of the experiments can be found at \texttt{https://youtu.be/OQiLbEaZsZw}.}
\label{fig:case2}
\end{figure*}
Similar to the previous case, in Fig.~\ref{fig:case2} there are four snapshots from the video of the Robotarium experiments. The robots drive from their initial positions (Fig.~\ref{subfig:case2:a}) to a \textit{centroidal Voronoi configuration} (Fig.~\ref{subfig:case2:b}), i.\,e. a minimum of the cost $J_1$ \cite{cortes2004coverage}. At this point, the global task specification $\pi^*$ is switched from $\pi^*=[1,0]\tr$ (surveillance) to $\pi^*=[0,1]\tr$ (formation control). From this time on, the solution of the optimization problem is such that the robots change priority to perform formation control and the input $u$ drives them to the locations indicated by the red dots (Fig.~\ref{subfig:case2:c}~and~\ref{subfig:case2:d}).

The experiments demonstrate how the constraint-based optimization formalism presented in this paper can  allow robots to dynamically prioritize between different tasks taking into account the robot heterogeneity and global task allocation specifications. The QP in \eqref{eq:qprelax} can be efficiently solved to simultaneously obtain the choice of which task to execute with highest priority, encoded by the values of $\alpha$ and $\delta$, and the control input $u$ that leads to the actual execution of tasks.

\section{Conclusions} \label{sec:conc}
In this paper, we presented a task allocation algorithm which optimally assigns task priorities to a team of robots with heterogeneous task capabilities. Using results from our previous work on constraint-based task execution, we allow robots to prioritize different tasks by adding relaxation variables in the constraints corresponding to the different tasks. This leads to a dynamic task allocation algorithm specially tailored for long-term autonomy applications which explicitly accounts for the heterogeneity in the robot capabilities as well as global specifications on the task allocation. The algorithm has been formulated as a quadratic program which can be quickly and efficiently solved by each robot. Its solution provides the robots both with the task priorities and the control inputs required to execute the prioritized tasks. Multiple robot experiments showcased the main features of the task allocation algorithm, such as its convergence properties and its efficient exploitation of robot heterogeneity.

\bibliographystyle{IEEEtran}
\bibliography{bib/IEEEabrv,bib/ecc2019ref}

\end{document}